\PassOptionsToPackage{dvipsnames,table}{xcolor}
\documentclass[10pt,twocolumn,letterpaper]{article}
\usepackage{times}
\usepackage{epsfig}
\usepackage{graphicx}
\usepackage{amsmath}
\usepackage{amssymb}
\usepackage{booktabs}
\usepackage{tikz}
\usepackage{comment}
\usepackage{amsmath,amssymb} 
\usepackage{color}
\usepackage{cite}
\usepackage[font=small,labelfont=bf,tableposition=top]{caption}
\usepackage{times}
\usepackage{epsfig}
\usepackage{graphicx}
\usepackage{amsfonts}
\usepackage{bbm}
\usepackage{pifont}
\usepackage{breakurl}  
\usepackage{parskip}

\usepackage[pagenumbers]{cvpr} 

\usepackage{epsfig}
\usepackage{graphicx}
\usepackage{amsmath}
\usepackage{amssymb}
\usepackage{enumitem}
\usepackage{url}
\usepackage{float}
\usepackage{xcolor,colortbl}
\definecolor{Gray}{gray}{0.85}

\usepackage{multirow}

\usepackage{booktabs} 
\usepackage{subcaption}
\usepackage{esint}

\newcommand{\puntawat}[1]{{\color{magenta} {[Puntawat: #1]}}}

\newcommand{\vect}[1]{\ensuremath{\mathbf{#1}}}
\renewcommand{\P}{\mathbf{P}}

\newcommand{\cmark}{\ding{51}}%
\newcommand{\xmark}{-}

\usepackage[pagebackref=true,breaklinks=true,letterpaper=true,colorlinks,bookmarks=false]{hyperref}
\def\paperID{*****} 
\def\confName{CVPR}
\def\confYear{2025}

\title{Where Is The Ball: 3D Ball Trajectory Estimation From 2D Monocular Tracking}

\author{Puntawat Ponglertnapakorn \quad Supasorn Suwajanakorn\\
VISTEC\\
Rayong, Thailand\\
{\tt\small \{puntawat.p\_s19, supasorn.s\}@vistec.ac.th}
}

\begin{document}

\thispagestyle{empty}


\twocolumn[{%
\renewcommand\twocolumn[1][]{#1}%
\maketitle

}]

  \begin{abstract}
We present a method for 3D ball trajectory estimation from a 2D tracking sequence. To overcome the ambiguity in 3D from 2D estimation, we design an LSTM-based pipeline that 
utilizes a novel canonical 3D representation that is independent of the camera's location to handle arbitrary views and a series of intermediate representations that encourage crucial invariance and reprojection consistency. 
We evaluated our method on four synthetic and three real datasets 
and conducted extensive ablation studies on our design choices.
Despite training solely on simulated data, our method achieves state-of-the-art performance and can generalize to real-world scenarios with multiple trajectories, opening up a range of applications in sport analysis and virtual replay. 
Please visit our page: \textit{\textcolor{pink}{\url{https://where-is-the-ball.github.io/}}}.

\end{abstract}
\vspace{-0.3cm}
  \vspace{-1em}
\section{Introduction}
A ball bouncing is a familiar sight to humans from a very young age. The setup is simple and its physics well-understood: curved trajectory, bouncing, gravity, and momentum. Despite its simplicity, this setting is the basis for a large number of sports and recreational activities. The ability to reconstruct a ball's trajectory can provide further insights and understanding for those activities as well as enable important applications, such as post-match sports analysis or immersive virtual replays. However, the vast majority of such content is in the form of monocular videos, and determining the exact 3D location at any given time from such 2D input remains challenging.

\begin{figure}
\centering
  \includegraphics[width=\linewidth]{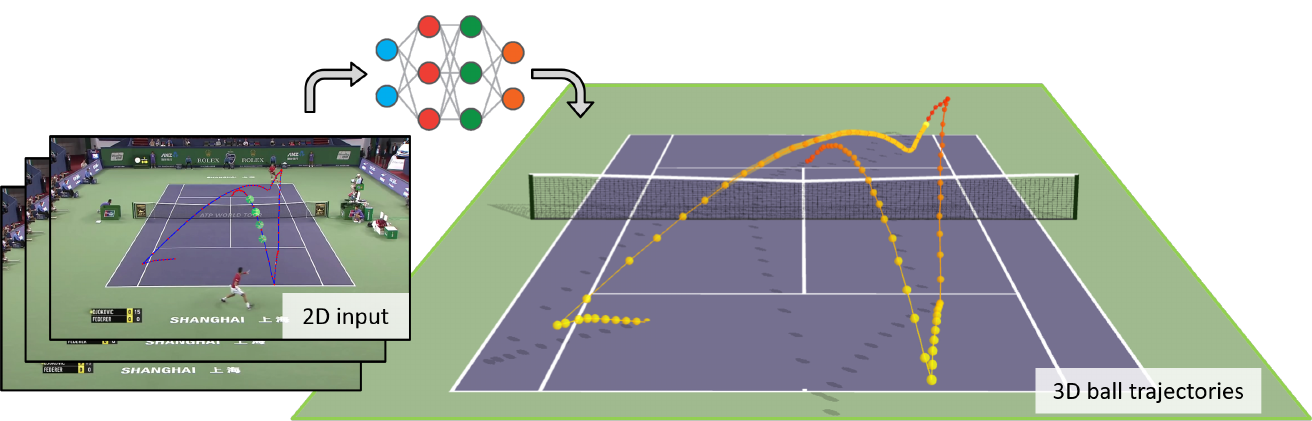}
  \vspace{-0.65cm}
  \caption{Given a 2D ball tracking sequence, we estimate the ball's 3D motion, which includes multiple bounces and hits.}
  
  \vspace{-1.5em}
  \label{ray_plane}
\end{figure}

\vspace{-0.5em}
The main difficulty arises from the inherent ambiguity where a 2D trajectory input can have multiple valid 3D motions that project to the same 2D trajectory.
Past solutions rely on various cues, such as the shadow of the ball, or estimating the ball size in pixel and relating it to the distance.
These geometric-based approaches often have restricting requirements and are not applicable to the vast majority of existing videos. 
Physics-based techniques place strong assumptions on the ball motion and require the entire trajectory to be precisely segmented into multiple projectiles, where each can be modeled with a simple physics equation. In practice, such segmentation is highly error-prone and sensitive to tracking noise. 
Learning-based approaches, on the other hand, attempt to learn physical motion priors from data; however, none has demonstrated a working solution that solves real-world bouncing balls with multiple continuous trajectories.

\vspace{-0.5em}
One major obstacle for learning-based techniques is the lack of large real-world training data with 3D ground truth, which is often difficult to collect. A common solution is to learn instead from simulated data; however, even with unlimited data, simply treating this problem as sequence modeling and directly regressing 2D to 3D coordinates often generalizes poorly to real-world scenarios. This can happen when crucial properties such as invariance to the camera intrinsic, shifting heights, or geometric-based reprojection consistency are ignored. Here we demonstrate that the key ingredient to our state-of-the-art performance is the right motion representations that leverage these inductive biases in our learning-based pipeline.

\vspace{-0.5em}
In particular, we designed a novel canonical representation that is independent of the camera parameters to handle multiple input viewpoints and a series of intermediate representations that successively refine the output trajectory by exploiting advantages of both relative and absolute coordinates. With these representations, our pipeline based on simple homogeneous LSTMs significantly outperforms other competing techniques
and can generalize to real-world trajectories despite training from simulation.
In summary, our contributions are:
\begin{itemize}[noitemsep,topsep=0pt]
    \item A state-of-the-art pipeline for estimating 3D trajectory of a bouncing ball from a sequence of 2D positions. This pipeline can handle real-world scenarios with multiple continuous trajectories, not demonstrated by prior work.
    \item Novel representations that support training and inference on multiple camera viewpoints within a single network.
    \item An extensive analysis of different trajectory parameterizations and our architectural design.
    \item A dataset of a real bouncing ball with 3D ground truth, which will be released along with our source code.
\end{itemize}

  \section{Related Work}

Our problem setup has been mostly explored in the area of sport video analysis. The accurate estimation of a ball trajectory in competitive sports, such as soccer, basketball, etc. is essential for the game understanding. Modern systems, such as the Goal-Line Technology~\cite{fifa}, provide necessary information for the development of the game in real time or for the judge through automatic line calling~\cite{hawkeye}, while other frameworks~\cite{second_spectrum} provide statistics for after-game analysis. However, most of these commercial products require expensive and elaborated multi-view setups such as Intel's True View~\cite{intel_true_view} or special tracking devices.

\vspace{-0.5em}
Techniques for estimating the 3D trajectory of a ball in motion from 2D input can be categorized by the type of input capture. Many techniques rely on a calibrated multi-camera setup and solve the 3D reconstruction by detecting the ball across all views and performing triangulation \cite{kamble2019convolutional,kim2019precise,ren2010multi,park20153d,kumar20113d,miyata2017ball}, while some others use stereo cameras \cite{liu2014improved,zhang2010visual,birbach2009multiple}. Our work focuses on a setup with a fixed, monocular video capture of the ball, where triangulation-based techniques are not applicable.

\vspace{-0.5em}
For monocular video setups, inferring the 3D location of a ball from 2D pixels is inherently ill-posed, despite the geometric/appearance cues that often appear in the video frames.
%
Reid et al.~\cite{reid19983d} utilize the shadow from the ball and a reference player to obtain the ball's height, but not all illumination or weather conditions can produce sufficient shadows.
Calandre et al. \cite{calandre2021extraction} estimate the size of a ping pong ball in pixels and then convert it into the distance from the camera using calibrated camera parameters. However, in sports such as tennis and soccer, the distance between the camera and the ball is too large for accurate size estimation. Additionally, in these sports, the ball typically moves very fast, which may introduce motion blur and severely degrade performance.
Mocanu et al. \cite{mocanu2017estimating} propose a learning method that uses energy-based restricted Boltzmann machines to predict the 3D ball position from its 2D projection. However, the underlying learning algorithm is complex and difficult to train~\cite{fischer2014training,romero2019weighted}. 

\vspace{-0.5em}
Since the physics of ball motion is well understood, many methods~\cite{yamada2002tracking,ribnick2008estimating,ohno2000tracking,BMVC.25.43,chen2009physics,shen20163d} incorporate physical constraints to compensate for the lack of 3D information and inconsistent 2D observations. The key idea is to estimate the physical parameters that best explain the detected trajectory, such as velocity and initial force. However, these methods require segmenting the input trajectory into individual projectiles or linear motions, a process highly prone to errors. There are heuristics that can be used for trajectory segmentation, such as detecting velocity changes, but in real-world scenarios, noisy and missing ball detections render those methods impractical. Chen et al. \cite{chen20113d} classify each trajectory segment as a pass (linear motion) or a cross (parabola motion) but require a set of hard-coded rules that do not generalize to more complex motions.
Other methods impose additional constraints, such as assuming that projectile motion occurs within a vertical 2D plane~\cite{kim1998physics,ren2004general,metzler20133d}. This assumption can break down for curved trajectories caused by lateral motion or spin. Our data-driven approach avoids such assumptions and can, in principle, learn any trajectory pattern given appropriate training data.

\vspace{-0.5em}
A number of studies focus on analyzing the dynamics of moving objects in an environment. For example, \cite{mottaghi2016newtonian} aims to learn the dynamics of an object given a single image, \cite{innamorati2019neural,purushwalkam2019bounce} generate plausible trajectories of virtual objects as they interact with an environment estimated from a still image, \cite{MonszpartThuereyMitra:SMASH:2016} reconstructs the 3D trajectories of colliding objects from a video, \cite{rozumnyi2021shape, rozumnyi2022motion} recover 6-DoF pose and shape of a fast moving object from motion-blurred images, and \cite{bhat2020adabins} estimates the physical parameters from a free flight video. However, those tasks are different from ours as we focus on estimating the exact 3D position of the ball given its projection from a 2D tracking sequence.

\vspace{-0.5em}
Recent work, SynthNet \cite{ertner2024synthnet}, proposes a two-stage pipeline incorporating tennis physics: first detecting ball hits and bounces to segment trajectories, then reconstructing the corresponding 3D trajectories by predicting initial conditions of projectile motion. 
While their pipeline enforces physical constraints specific to tennis ball dynamics, it does not enforce projection consistency. As a result, errors in estimating the initial conditions can significantly degrade the quality of the reconstruction. In contrast, our method implicitly learns the ball's physical motion from simulated data and directly predicts the height corresponding to each 2D tracking point, inherently ensuring that the reconstructed 3D trajectory always aligns with the original 2D input.
A similar combination of physics with learning-based methods has also been applied to other research directions, such as human motion estimation~\cite{RempeContactDynamics2020,Vondrak_phy_mot}, human pose tracking~\cite{brubaker_kneed,brubaker_anthropo}, and human-object-scene interactions~\cite{Hodgins:2017:DOE,Zhu2015UnderstandingTT,Zhu_inferring_force}. However, these methods are human centered and not directly applicable to more general settings.

  \begin{figure*}
\centering
  \includegraphics[width=\textwidth]{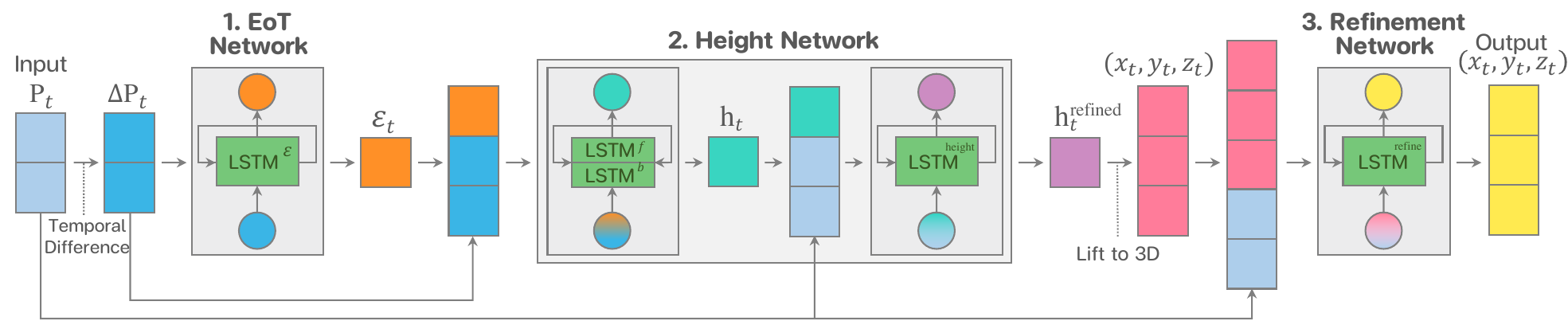}
  \vspace{-0.65cm}
  \caption{\textbf{Method overview.} Given a 2D ball tracking sequence $(u_t, v_t)$, we first convert each tracked point to our novel 3D plane points parameterization $\P = (\mathbf{p}_\text{ground}, \mathbf{p}_\text{vertical})$ and then predict the 3D ball coordinates $(x_t, y_t, z_t)$. Our pipeline consists of 3 main components. 1) EoT network takes in the plane point temporal differences and predicts the end-of-trajectory (EoT) probability. 2) Height network takes in the EoT probability and plane points to predict the height, which is then converted to a 3D coordinate. 3) Refinement network then refines the coarse 3D coordinate and produces the final output.
  }
  \label{pipeline}
  \vspace{-0.6cm}
\end{figure*}

\begin{figure}
\centering
  \includegraphics[scale=0.57]{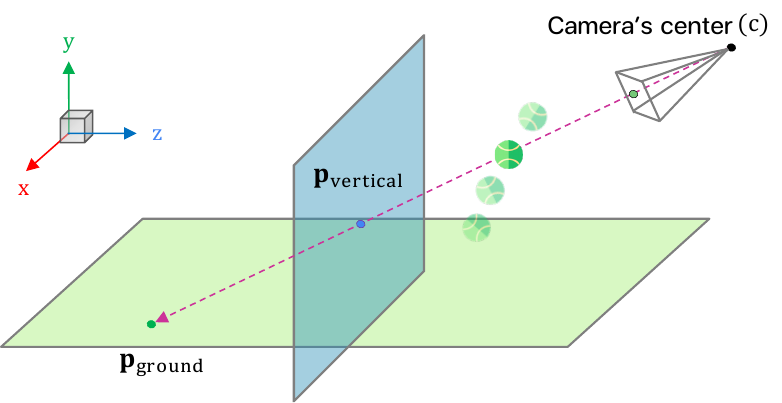}
  \vspace{-0.2cm}
  \caption{\textbf{Ray parameterization.} We represent a 2D track point as the associated 3D viewing ray, parameterized as two intersection points $\mathbf{p}_\text{ground}, \mathbf{p}_\text{vertical}$ of the ray with the ground plane (y=0) and a vertical plane (e.g., z=0).}
  \label{ray_plane}
  \vspace{-1.6em}
\end{figure}

\section{Approach}\label{approach}

Given a 2D ball tracking sequence, our goal is to estimate the corresponding 3D position of each 2D point. We focus on real-world bouncing scenarios across different sports. Each input sequence may contain multiple trajectories, each beginning when a force is applied to the ball (e.g., a soccer player's kick) and ending just before another force acts or the ball comes to rest. A single trajectory can include multiple bounces on the ground. 
We assume that both the beginning and end of the input sequence lie on the ground ($y = 0$) and the camera parameters are known.

\vspace{-0.5em}
To facilitate 3D predictions from different camera perspectives, one of our key ideas is to first map the raw 2D track points into a canonical 3D space. In particular, we represent each 2D track point using a 3D ray and parameterize it further as intersection points on two perpendicular planes. Secondly, we observe that absolute and relative coordinates have their unique advantages and design a pipeline that exploits both using multiple types of parameterization, which is proved much more effective than na\"ive approaches in Section \ref{sec:input_param}.
\subsection{Trajectory estimation pipeline}\label{sec:pipeline}
Our pipeline consists of three main components (Figure~\ref{pipeline}): 1) An end-of-trajectory prediction network, which predicts the trajectory boundaries of the input sequence by predicting the probability of the ball ending its current trajectory for each time step, 2) A height prediction network, which takes the input sequence and the end-of-trajectory probabilities to estimate the height from the ground for each time step, 3) A refinement network, which refines the 3D coordinates reconstructed from the predicted heights. The input to these components will be our new representation of the 2D track points, which will be explained next.

\vspace{-1em}
\subsubsection{Input parameterization}
To handle 2D tracking inputs that may come from different cameras from various locations, we propose to reparameterize each 2D point in the input sequence as a representation in a canonical 3D space that is independent of the camera's parameters, such as its location, orientation, or focal length. 
That is, instead of using 2D coordinates directly as input to our prediction networks, we first back-project each 2D pixel to its corresponding 3D viewing ray $\mathbf{r}(s) = \mathbf{c} + \mathbf{d}s$ that starts from the camera's center of projection $\mathbf{c} \in \mathbb{R}^3$ and points toward the pixel on the image plane in the direction $\mathbf{d} \in \mathbb{R}^3$. Given a 2D pixel location $(u, v)$ and the extrinsic $E \in \mathbb{SE}(3) \subset \mathbb{R}^{4 \times 4}$, the center and direction can be computed by:\vspace{-0.7em}
\begin{align}
\mathbf{c} &= \psi(E^{-1}\begin{bmatrix}0, 0, 0, 1\end{bmatrix}^\top)\\
\mathbf{d} &= \psi(E^{-1}\begin{bmatrix}u-p_x, v-p_y, f, 0\end{bmatrix}^\top)
\end{align}
where $\psi:\mathbb{R}^4\rightarrow \mathbb{R}^3$ is the dehomogenize operator that removes the last element: $\psi([x\ y\ z\ w]) = [x\ y\ z]$, $f$ is the focal length, and $(p_x, p_y)$ is the principle point.

This ray representation is not unique, and ideally we want all collinear rays to share the same representation. We solve this by reparameterizing this ray as two intersection points of the ray with two planes: the ground plane ($y=0$) and a vertical plane (e.g., $z=0$). In sport applications such as tennis, this vertical plane can be set coplanar with the court's net, as a convenient choice (see Figure \ref{ray_plane}). Our input representation for each 2D track point is thus given by $\mathbf{P} = (\mathbf{p}_\text{ground}, \mathbf{p}_\text{vertical})$, where $\mathbf{p}_\text{ground}, \mathbf{p}_\text{vertical} \in \mathbb{R}^3$ are the intersection points of the ray with the two planes. Here we assume that the camera is placed high enough in the scene and facing downward so that no rays are parallel to the ground or the vertical plane. In practice, we drop the y-coordinate from $\mathbf{p}_\text{ground}$ and z-coordinate from $\mathbf{p}_\text{vertical}$ because they are always zero, resulting in $\P \in \mathbb{R}^{4}$. 

\vspace{-1em}
\subsubsection{End-of-trajectory (EoT) prediction network}
\label{sec:eot}
The goal of this network is to predict the trajectory boundaries of the input sequence. 
Our intuition is that if our height prediction network has some information about when the trajectories start and end, or when the ball changes its course, it could solve a simpler estimation problem involving one trajectory moving in a relatively constant x-z direction. Unlike in physics-based approaches, this information is not used to hard-segment trajectories but will be used as an auxiliary signal for the height prediction network.

\vspace{-0.3em}
Specifically, the EoT network takes as input the temporal differences of plane points $({\Delta{\P}_{t}}) = (\P_{t+1}-\P_t)$ and predicts an EoT probability $\varepsilon_t \in [0, 1]$ of the ball ending its current trajectory or coming to a stop.
This ``relative'' representation helps provide invariance to the initial plane point locations. In other words, if a network with this invariance learns to predict a sequence starting at $\P_1$, it can predict shifted versions of the same sequence that start at $\P_1 + (\vect{a}, \vect{b})$ for any $a, b \in \mathbb{R}^2$.
This network ($\text{LSTM}^\varepsilon$) is modeled with a stack of 3 bidirectional-LSTMs with shortcut connections between each layer inspired by \cite{yu2017improved}. The last hidden state is connected to 3 fully-connected layers to output the EoT probability $\varepsilon_t \in [0, 1]$. 
The architecture details are provided in Appendix \ref{app:net_arch}.
\vspace{-1em}
\subsubsection{Height prediction network}
\label{sec:height}
For our problem, predicting 3D coordinates directly from the input sequence is one possible design choice; however, this tends to perform poorly, and the prediction simply ignores important projection consistency (i.e., the projection of the predicted 3D point should match the pixel). Predicting depth values can ensure this consistency, though depths are defined with respect to the arbitrary camera's location, which further complicates the prediction. Instead, we first predict the height of each point, which has only one degree of freedom and is independent of the camera's location. With our assumption that no rays are parallel to the ground, a height $h$ of a track point will uniquely determine its 3D coordinate $\vect{r}(s^*)$, where $s^*$ is the solution to $\vect{r}^y(s^*) = h$ and $\vect{r}^y$ is the y-coordinate of the viewing ray associated with the track point. Thus, the predicted height is always projection-consistent and will be converted to 3D coordinates later (Section~\ref{sec:refine}).

\vspace{-0.5em}
To predict the height, we first use two unidirectional LSTMs ($\text{LSTM}^{f}$, $\text{LSTM}^{b}$) that first compute forward and backward temporal height differences. Then, we aggregate and combine them to produce a single height sequence $h_t$, which will be refined with another bidirectional $\text{LSTM}^\text{height}$ to produce $h_t^\text{refined}$. 
Specifically, $\text{LSTM}^{f}$ takes as input $({\Delta\P_{t}}, \varepsilon_t, h_t^f)$ at each time $t$ and predicts $\Delta h_t^f$, where $h_t^f$ is computed by accumulating $\Delta h_{t-1}^f$ from the earlier step: $h_t^f = h_{t-1}^f + \Delta h_{t-1}^f$ and $h_0^f=0$. The backward $\text{LSTM}^{b}$ works similarly but starts accumulating from $h_N^b=0$ backward. This step yields $h_t^f$ and $h_t^b$ through accumulation, then we combine them with a simple ramp sum:
\begin{align}
    h_t &=(1-w_t)h_t^f + (w_t)h_t^b
    \label{eq:ramp}
\end{align}
where $w_t = (t-1)/(N-1)$. The motivation for combining both directions is to reduce long aggregation errors by relying more on the forward sum near the beginning and the backward sum near the end. Finally, the weighted sum $h_t$ together with the plane points input $(h_t, \P_{t})$ will be fed to $\text{LSTM}^\text{height}$ to predict $h_t^\text{refined}$. The architecture of $\text{LSTM}^\text{height}$ is identical to $\text{LSTM}^\varepsilon$ (Sec. \ref{sec:eot}).

\vspace{-0.3em}
This design makes use of the relative representations for both the input and output of $\text{LSTM}^{f}$ and $\text{LSTM}^{b}$, which help achieve invariance to shifting heights. However, the relative representations alone lack the awareness of absolute positioning and can drift over time or protrude underground. The use of $\text{LSTM}^\text{height}$ here helps alleviate this issue by operating on absolute heights and absolute plane points to refine the height sequence.

\vspace{-1em}
\subsubsection{Refinement network} 
\label{sec:refine}
The earlier height parameterization ensures that the projections of the predicted points always match their corresponding pixels. However, in real-world uses, the input 2D tracking sequence may come from a tracking algorithm, which can give noisy 2D estimates. Constraining the projected 3D coordinates to exactly match these noisy 2D estimates will subsequently lead to the wrong 3D predictions. So in this step, we use another network to refine the prediction by giving it the flexibility to modify the actual 3D coordinates. This also helps the network utilize other priors such as trajectory smoothness in 3D space learned from simulation.

\vspace{-0.5em}
Given our refined height $h_t^\text{refined}$, we convert it to the corresponding 3D coordinate $\vect{r}_t(s_t^*) = (x_t, y_t, z_t)$, again via solving $\vect{r}_t^y(s_t^*) = h_t^\text{refined}$. The resulting 3D sequence together with the plane points $(x_t, y_t, z_t, \vect{P}_t)$ will be input to our refinement network to predict $(\delta x_t, \delta y_t, \delta z_t)$ and output the final 3D coordinates $(x_t, y_t, z_t)^\text{final}=(x_t + \delta x_t, y_t + \delta y_t, z_t + \delta z_t)$. The choice of predicting the deltas makes it easier to initially learn the identity function and focus on the refinement, as motivated in ResNet \cite{he2016deep}. This refinement network ($\text{LSTM}^\text{refine}$) is also a stack of 3 BiLSTMs similar to $\text{LSTM}^\varepsilon$ and $\text{LSTM}^\text{height}$.

\subsection{Network training}\label{sec:training}
We train our networks using simulated data generated from the PhysX physics engine\cite{unity} in Unity. We simulate a bouncing ball by applying a series of impulse forces and record the 3D positions, 2D projected coordinates, and end-of-trajectory flags for each time step. These variables will be used as supervised training data. 
Training, validation, and testing data are generated separately.
We use the following loss functions to jointly train all networks.

\vspace{-0.5em}
\textbf{End-of-Trajectory loss.} We use weighted binary cross-entropy for the EoT prediction network:
\vspace{-0.2em}
\begin{align}
\resizebox{1\hsize}{!}{$
\mathcal{L}_{\varepsilon} = - \frac{1}{N}\sum_{t=1}^{N} \left({\gamma}\varepsilon_t^\text{gt}\log(\varepsilon_t) + (1-\gamma)(1-\varepsilon_t^\text{gt}) \log (1-\varepsilon_t)\right)$}
\end{align}

\vspace{-1em}
where $\varepsilon_t^\text{gt}$ is the ground-truth EoT binary flag, $\varepsilon_t$ is our predicted EoT probability, and $\gamma$ is a balancing weight between the two classes ($\varepsilon_t=0$ or 1).

\vspace{-0.5em}
\textbf{3D reconstruction loss.} We use the L2 loss for the reconstruction error of 3D coordinates:
\vspace{-0.2em}
\begin{align}
\mathcal{L}_\text{3D} = \frac{1}{N} \sum_{t=1}^{N}\left\| (x_t, y_t, z_t)^\text{gt} - (x_t, y_t, z_t)^\text{final}\right\|^2_2
\end{align}
where $(x_t, y_t, z_t)^\text{gt}$ is the ground-truth 3D coordinate, and $(x_t, y_t, z_t)^\text{final}$ is the predicted coordinate after refinement.


\vspace{-0.5em}
\textbf{Below ground loss.} We penalize every point below the ground plane $(y=0)$ using its squared distance:
\vspace{-0.2em}
\begin{align}
\mathcal{L}_{B} = \frac{1}{|\mathbb{Y}|}\sum_{y \in \mathbb{Y}} y^2
\end{align}
where $\mathbb{Y} = \{y_t^\text{final}\ |\ y_t^\text{final} < 0\}$ contains the y coordinates of all the predicted points below the ground.

\vspace{-0.5em}
\textbf{Total Loss.}
We optimize all LSTMs together using the sum of all loss functions:
\vspace{-0.2em}
\begin{align}\label{eq:totalloss}
\mathcal{L}_\text{Total} = \lambda_{\varepsilon}\mathcal{L}_{\varepsilon} + \lambda_\text{3D}\mathcal{L}_\text{3D} + \lambda_{B}\mathcal{L}_{B}
\end{align}

\vspace{-1em}
where $\lambda_{\varepsilon}, \lambda_\text{3D}$ and $\lambda_{B}$ are balancing weights.
\section{Experiments}
\label{section:exp}
We conduct experiments on four synthetic and three real-world datasets, provide comparisons to state-of-the-art techniques, and perform ablation studies on parameterization, pipeline components, and loss functions. Implementation, simulation details, and runtime are in the Appendix.

\vspace{-0.5em}
\textbf{Evaluation metrics.} We use normalized root-mean-square error (NRMSE) for all experiments following \cite{mocanu2017estimating}, except when comparing with SynthNet \cite{ertner2024synthnet} (Section~\ref{sec:compare_synthnet}), where we follow their evaluation protocol.
NRMSEs are RMSEs computed between predicted and ground-truth coordinates, normalized by the maximum range in the $x$, $y$, or $z$ dimensions of the ground truth in each dataset. As acceptable error is application-specific, we also report NRMSEs relative to camera distance and area width in Appendix \ref{app:more_results}.

\subsection{Datasets}
This section describes all datasets used in our experiments, all with 3D ground truth except Real TrackNet~\cite{huangtracknet}, used for comparison with SynthNet~\cite{ertner2024synthnet} (Section \ref{sec:compare_synthnet}).

\vspace{-0.5em}
\textbf{Real TrackNet:} This dataset \cite{huangtracknet} contains 81 video clips from 10 tennis matches, captured from a broadcast camera, along with 2D ball-tracking annotations. These clips have varying numbers of strokes between 1-10. We calibrated the camera through solving the perspective-n-point problem with 2D-3D correspondence provided by a court detection algorithm~\cite{farin_court}.

\vspace{-0.5em}
Next, we describe the datasets used for comparison with other baseline methods (Section \ref{comp_single_laucnh}) and ablation studies. 

\vspace{-0.5em}
\textbf{Real Mocap:}  
We captured a ping pong ball’s bouncing motion in our $\approx10^2$m$^2$ motion capture studio using an IR reflective sticker and eight synchronized IR cameras at 50 fps. The dataset consists of 344 sequences (103,872 data points) with highly accurate 3D ground truth from the Mocap system, and 2D trajectories from all eight cameras.

\vspace{-0.5em}
\textbf{Real IPL:} This dataset \cite{fotouhi2017projection} contains a 2-minute capture of a real soccer match from 6 synchronized cameras on both touchlines of the pitch and 2D tracking annotations. The 3D ground truth was estimated using triangulation and the camera pose estimation pipeline in \cite{rematas2018soccer}. Nine sequences were successfully calibrated, with missing track points filled using an autoregressive LSTM (detailed in our Appendix).

\vspace{-0.5em}
\textbf{Synthetic TrackNet / Mocap / IPL datasets:} We created synthetic counterparts for each real dataset (Mocap, IPL, TrackNet) that match their camera parameters and trajectory characteristics. We simulate projectiles for Mocap and IPL and simulate a tennis game with two players for TrackNet. 
Each dataset contains 5,000 training sequences and 500 test sequences.

\vspace{-0.5em}
\textbf{Synthetic Single-Launch dataset:} To compare with \cite{mocanu2017estimating, shen20163d} and match their setups, we create this dataset with 300 training and 100 testing sequences of single-launch trajectories, where the ball is launched once and bounces until it stops. More details are in Appendix \ref{app:dataset_details}.

\subsection{Comparison with prior work} 
In this section, we compare our method with the most recent SOTA method, SynthNet \cite{ertner2024synthnet}, which is designed for and evaluated on tennis matches from the TrackNet~\cite{huangtracknet}. 
We also evaluate our method against existing methods \cite{mocanu2017estimating, shen20163d} that assume single-launch trajectories, a more restrictive setting than ours.
We exclude geometry-based approaches that rely on shadows or player height information \cite{reid19983d, kim1998physics}, as these assumptions are too restrictive and the required information is unavailable in our datasets and most real-world scenarios.
For completeness, we also compare \cite{mocanu2017estimating, shen20163d} on Real IPL and Mocap in Table \ref{sota_comp_real} (Appendix), where they produce large errors since these datasets contain trajectories with multiple launches beyond their assumptions\footnote{\label{fn:verification}\scriptsize 
Results were generated and tuned using their code, and verified with the authors.}

\vspace{-1.1em}
\subsubsection{Comparison on tennis matches---TrackNet \cite{huangtracknet}}
\label{sec:compare_synthnet}
We compare our method with SynthNet \cite{ertner2024synthnet} on Real TrackNet \cite{huangtracknet}. Following SynthNet's evaluation protocol, we assess the tennis ball's landing position using their proposed metrics: landing accuracy (T.F1 and T.acc) and landing error (LE). The contact point annotation frame is taken directly from TrackNet's labels and used to generate the 3D contact point ground truth via ray tracing from the calibrated camera onto the court surface. We present the results in Table~\ref{synthnet_tracknet}, with their results taken directly from their paper. Our method outperforms SynthNet across all metrics, achieving an average landing accuracy of 87.21$\%$, an F1-score of 0.807, and a significantly lower landing error of 0.63 meters compared to 3.58 meters for SynthNet.

\vspace{-1.1em}
\subsubsection{Comparison on single-launch trajectories}\label{comp_single_laucnh}
We evaluate against Shen et al.~\cite{shen20163d} and Mocanu et al.~\cite{mocanu2017estimating} on the Synthetic Single-Launch Trajectory dataset, which matches their problem setup and assumptions. 
The physics-based method~\cite{shen20163d} minimizes reprojection error by optimizing two physical parameters: initial velocity and position. It represents an improved physics-based method that builds upon \cite{ohno2000tracking, ribnick2008estimating} by incorporating a contact points constraint. Since no source code is available, we reimplemented this baseline. The learning-based method~\cite{mocanu2017estimating} models projectile motion using restricted Boltzmann machines. We train their model using their official code and train our method on the same 300 sequences. We use the same test set for all methods. Additionally, we assess robustness to tracking errors by testing with different levels of noise in the 2D input.

\begin{figure}
\centering
  \includegraphics[width=\columnwidth]{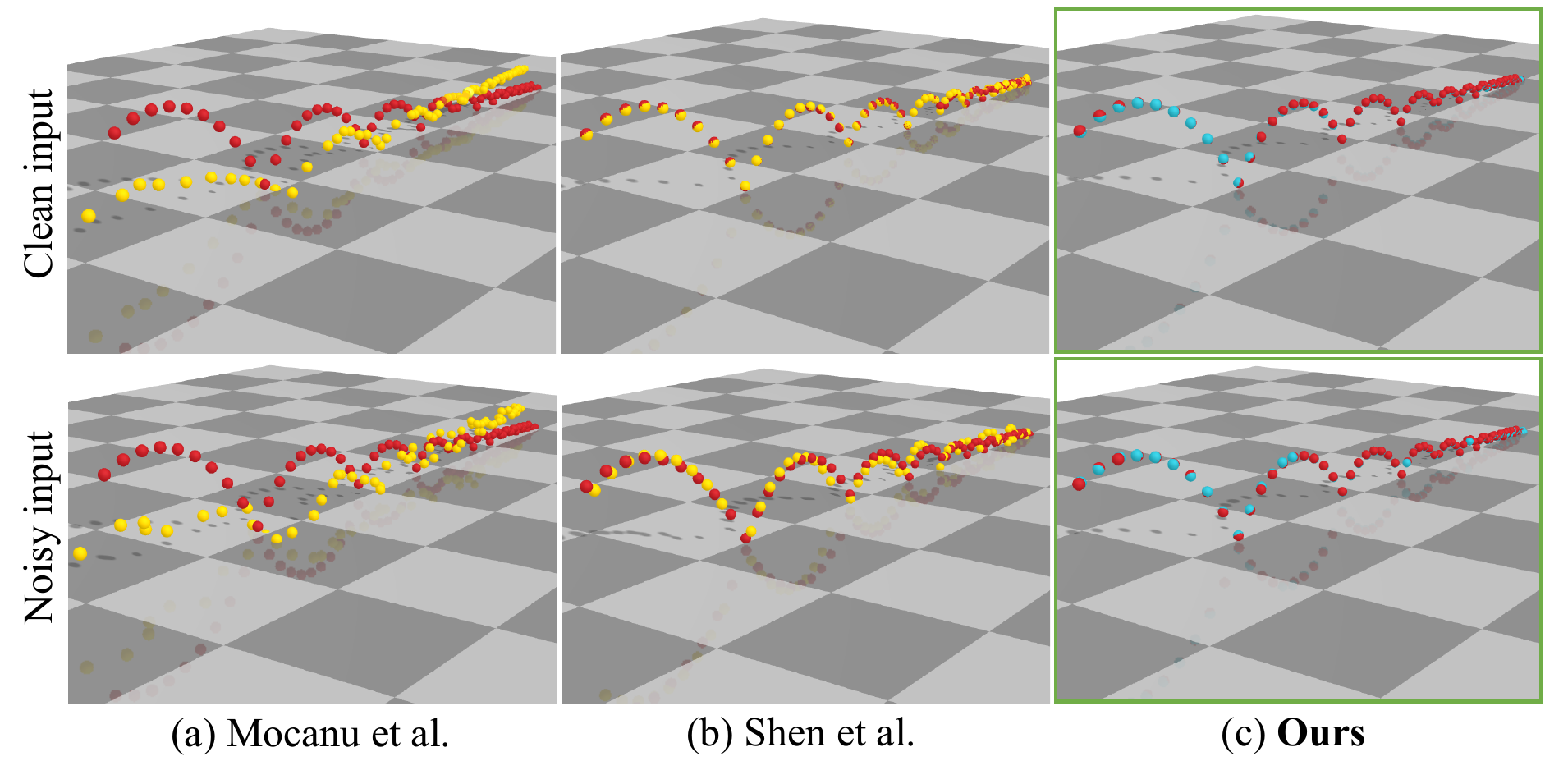}
  \vspace{-2em}
  \caption{Comparison with learning \cite{mocanu2017estimating} and physics-based \cite{shen20163d} methods. We add $\pm{25}$-pixel noise to the 2D input in the bottom row. Blue: ours. Yellow: prior work. Red: ground truth.
  }
  \label{fig_sota}
  \vspace{-1.7em}
\end{figure}

\vspace{-0.5em}
We report distance NRMSEs and standard errors in Table \ref{sota_table} and provide a qualitative comparison in Figure~\ref{fig_sota}. Our method achieves the best NRMSE of 0.03, outperforming \cite{mocanu2017estimating} (1.02)\footref{fn:verification} and \cite{shen20163d} (0.11). Our NRMSE translates to about 0.6cm RMSE on this dataset. Our method also degrades minimally compared to others when the noise increases up to 25 pixels (input resolution is 1664$\times$1088). 

\subsection{Ablation analysis}

\subsubsection{Input / output parameterization}
\label{sec:input_param}
For this ablation study, we evaluate our plane points parameterization against five other alternatives:
\setlist{nolistsep}
\begin{enumerate}
\vspace{-0.5em}
\item Pixel: uses 2D pixel coordinates as input.
\item Pixel + Extrinsic: uses 2D pixel coordinates concatenated with the flattened extrinsic $E \in \mathbb{SE}(3)$, which contains the camera's rotation and translation.
\item $\mathbf{p}_\text{ground}$ + ($\varphi_{az}, \theta_{el}$): uses our viewing ray technique but parameterizes the ray by the ground plane point and the azimuth and elevation angles in radian from the ground plane point to the camera center.
\item $\mathbf{p}_\text{ground}$ + ($\varphi_{az}^{\sin{},\cos{}}$, $\theta_{el}^{\sin{},\cos{}}$): is similar to (3.) except the azimuth and elevation are represented with the sine and cosine of their angles. 
\item $\mathbf{p}_\text{ground}$ + $\mathbf{p}_\text{vertical}$: is our proposed parameterization.
\end{enumerate}

We test each input parameterization in combination with two types of output parameterization: 1. predicting xyz directly and 2. predicting height (ours). Note that the two output types here refer to the parameterization \emph{before} the refinement step ($\text{LSTM}^\text{refine}$ is fixed and always refines 3D coordinates in all cases). 
The input/output dimensions of $\text{LSTM}^{\varepsilon, f, b, \text{height}}$ vary by parameterization, but the rest of the architectures remains the same. We use all three synthetic data and only two Real Mocap and IPL because Real Tennis lacks 3D ground truth for quantitative evalutions and consists of \emph{single-view} videos unfit for triangulation.

\begin{figure*}
\centering
  \includegraphics[width=1\textwidth]{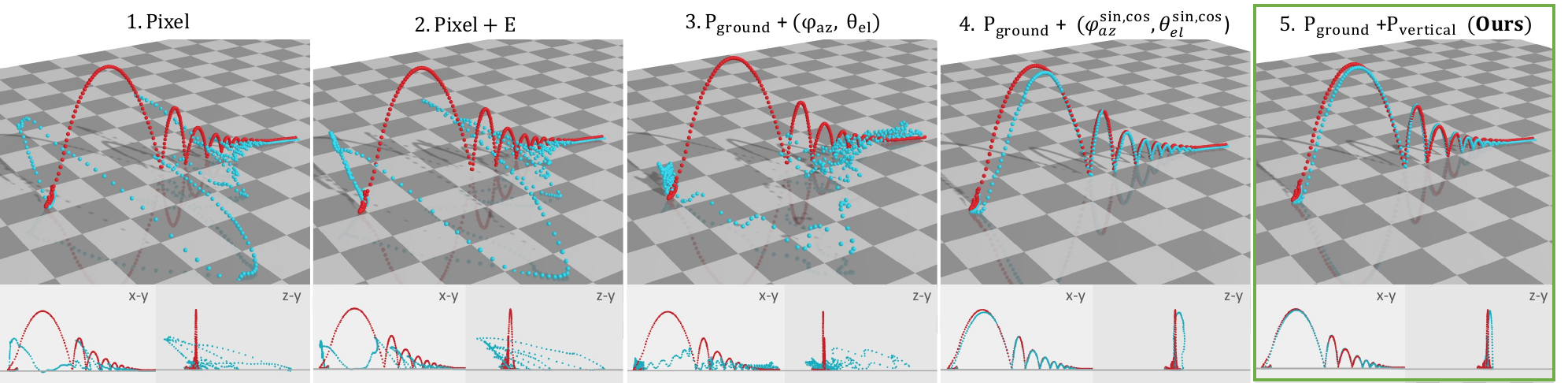}
  \vspace{-2em}
  \caption{\textbf{Different input/output parameterization types}. The predictions are in blue and ground truth in red. (y-axis points up)
  } 
  \label{fig_params}
  \vspace{-1em}
\end{figure*}

\vspace{-0.5em}
The results in Table \ref{features_table}
show that using the right parameterization is crucial and can produce significantly better results than the alternatives.
The na\"ive pixel parameterization performs poorly on every dataset even when the camera poses were provided. 
Our plane points parameterization for the viewing ray outperforms other ray parameterization, such as using the azimuth and elevation angles, and produces the lowest errors across all datasets. This could be due to the more direct correspondence between the motion on our vertical plane and the actual 3D motion (e.g., a 3D projectile directly shows up as a parabolic motion of $\vect{p}_\text{vertical}$), whereas the elevation or azimuth parameterization requires modeling complex and less-direct relationships between angles and 3D motion. 
For output parameterization, we found the na\"ive solution of directly predicting 3D coordinates highly prone to overfitting (i.e., the error gap between Real and Synthetic Mocap is very high). Figure \ref{fig_params} shows examples of the predicted trajectories from various parameterization.

\begingroup
\setlength{\tabcolsep}{1.5pt} 
\begin{table}[ht]
\centering
\vspace{-0.5em}
\caption{\textbf{Ablation study on input/output parameterization}. We report NRMSEs using different parameterization schemes. \emph{Output} parameterization is of the height network before refinement.}
\vspace{-0.8em}
\resizebox{\columnwidth}{!}{%
\begin{tabular}{lc|ccc|cc}
\toprule
\multicolumn{2}{c|}{\multirow{1}{*}{\textbf{Parameterization}}} & \multicolumn{3}{c|}{\textbf{Synthetic}} & \multicolumn{2}{c}{\textbf{Real}} \\
\multicolumn{1}{l}{Input} & \multicolumn{1}{c|}{Output}
& \multicolumn{1}{c}{Mocap}  & \multicolumn{1}{c}{Tennis} & \multicolumn{1}{c|}{IPL}
& \multicolumn{1}{c}{Mocap}  & \multicolumn{1}{c}{IPL} \\ \midrule

\multirow{2}{*}{Pixel} & xyz & 14.13 & 0.31 & 0.70 & 11.68 & 2.45 \\ 
{} & height & 0.08 & 0.30 & 0.05 & 4.63 & 2.63 \\ \cline{1-7}

\multirow{2}{*}{Pixel + $E$} \rule{0pt}{2.2ex} & xyz
& 6.29 & 0.66 & 0.69 & 34.64 & 2.21 \\
{} & height 
& 0.09 & 0.27 & 0.05 & 4.79 & 2.74 \\ \cline{1-7}

\multirow{2}{*}{$\mathbf{p}_\text{ground} + (\varphi_{az}, \theta_{el})$} \rule{0pt}{2.2ex} & xyz 
& 0.18 & 0.21 & 0.43 & 9.12 & 2.23 \\  
{} & height
& 0.10 & 0.19 & 0.02 & 4.70 & 2.47 \\  \cline{1-7}

\multirow{2}{*}{$\mathbf{p}_\text{ground} + (\varphi_{az}^{\sin{},\cos{}}, \theta_{el}^{\sin{},\cos{}})$} \rule{0pt}{2.2ex} & xyz
& 0.13 & 0.23 & 0.82 & 4.17 & 2.73 \\  
{} & height & 0.27 & 0.16 & 0.03 & 0.94 & 1.64 \\  \cline{1-7}

\multirow{2}{*}{$\mathbf{p}_\text{ground} + \mathbf{p}_\text{vertical}$\ \textbf{(Ours)}} \rule{0pt}{2.2ex} & xyz
& 0.11 & 0.24 & 0.37 & 0.85 & 2.81 \\
{} & height \textbf{(Ours)} & \textbf{0.05} & \textbf{0.09} & \textbf{0.01} & \textbf{0.68} & \textbf{0.74} \\ \bottomrule%
\end{tabular}}
\vspace{-0.5em}
\label{features_table}
\end{table}
\endgroup

\begin{figure*}
\centering
  \includegraphics[width=\textwidth]{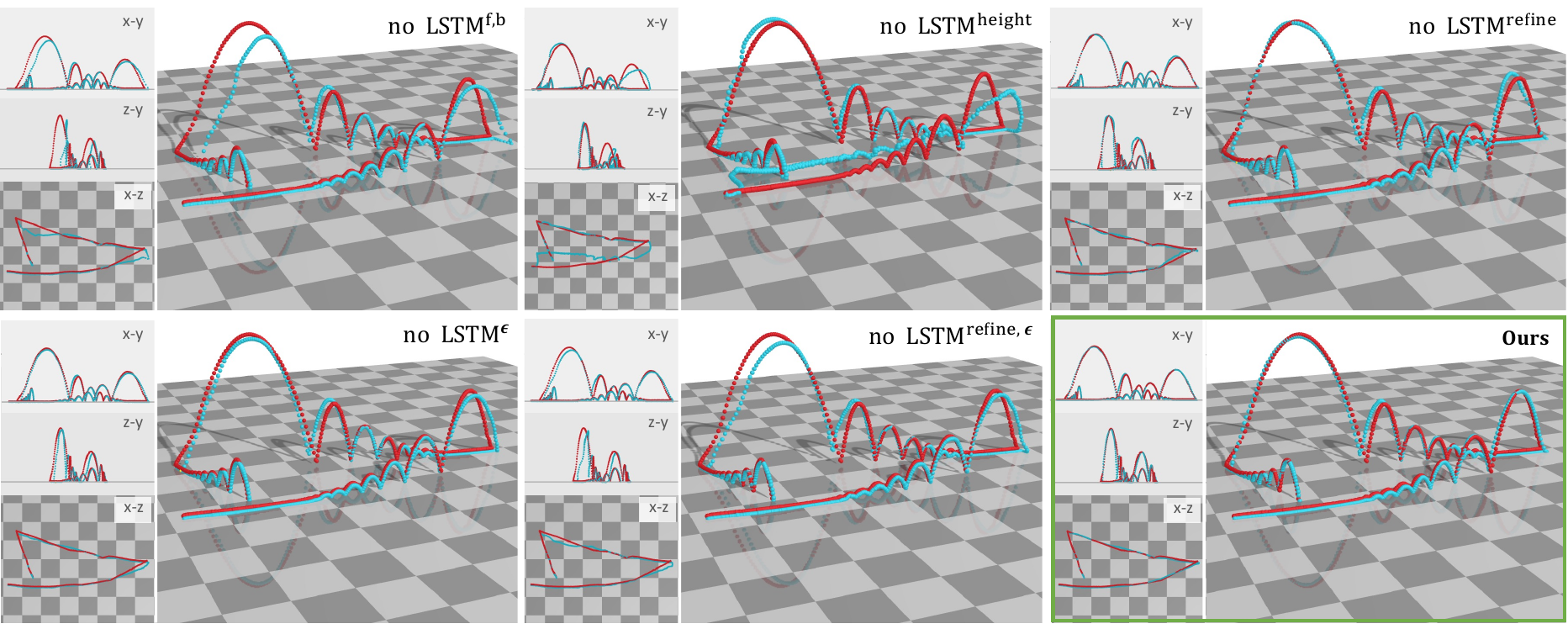}
  \vspace{-0.7cm}
  \caption{\textbf{LSTM ablation study}.
  The predictions are in blue and ground truth in red. (y-axis points up, each block is 50$\times$50 cm$^2$)
  }
    \vspace{-1.5em}
  \label{fig_pipe}
\end{figure*}

\vspace{-1em}
\subsubsection{Pipeline components}
Here we ablate LSTM components to study their contributions. We evaluate each variation on the same validation sets used in the earlier experiment and reported distance NRMSEs in Table \ref{module_table}.
The results show that the height and refinement networks ($\text{LSTM}^\text{height}$, $\text{LSTM}^\text{refine}$) are especially important, and without them the errors significantly increase across all datasets. Without the end-of-trajectory flags from $\text{LSTM}^\varepsilon$, the negative effect is large on Tennis and IPL, which tend to have less predictable forces from the players. This suggests that the EoT flags, which mostly indicate direction changes in the motion, can be more beneficial in such scenarios. By relying on the outputs of $\text{LSTM}^{f,b}$, $\text{LSTM}^\text{height}$ can utilize information from both forward and backward directions and prove helpful in reducing NRMSEs by about 6.2 (50cm) in Mocap and 0.9 (1m) in IPL.
The refinement network $\text{LSTM}^\text{refine}$ helps refine and smooth the trajectory output in 3D space as shown in Figure \ref{fig_pipe}.

\begin{figure*}[t]
\centering
  \includegraphics[width=0.84\textwidth]{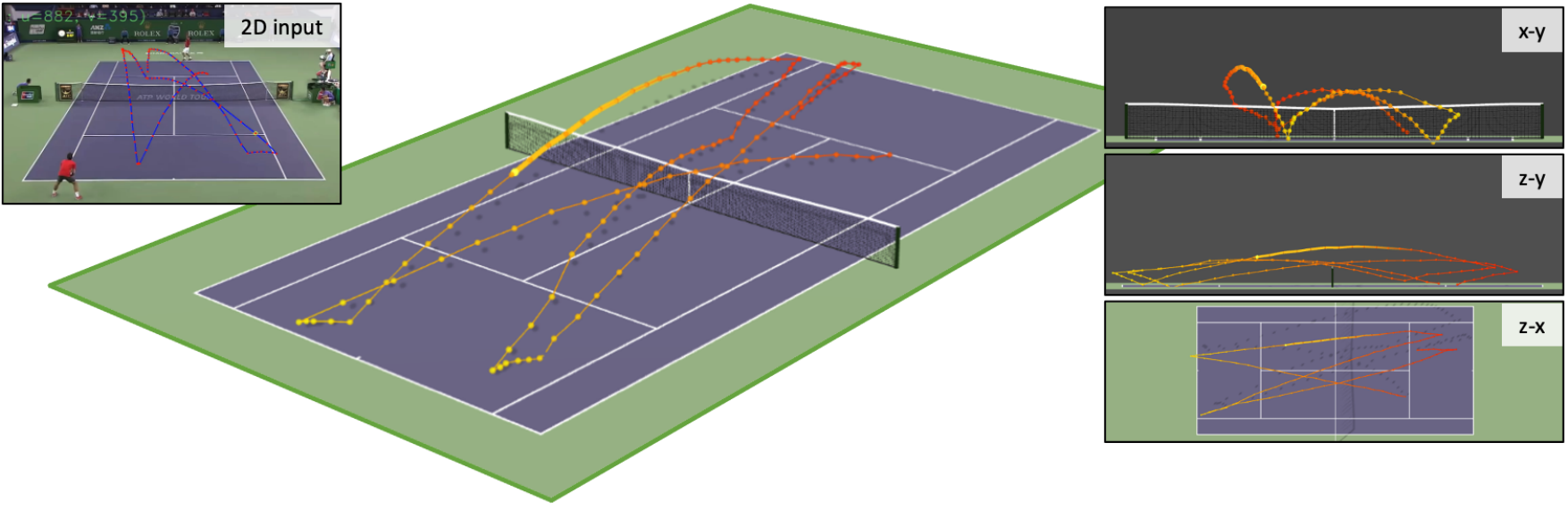}

  \vspace{-1em}
  \caption{From the 2D tracking of the tennis ball on the left, our method can successfully predict multiple consecutive 3D trajectories.
  } 
  \label{fig_tennis}
\end{figure*}

\vspace{-0.5em}
We show 3D visualizations of the output on a real tennis match in Figure \ref{fig_tennis} and Appendix \ref{app:more_results}. Our pipeline can predict challenging bounces that contain multiple back-and-forth hits by the two players in a tennis game.

\begingroup
\setlength{\tabcolsep}{5pt} 
\begin{table}[ht]
\centering
\caption{{\textbf{Ablation study on LSTM components}. We report NRMSEs using different configurations of components: $\varepsilon$ (Section \ref{sec:eot}), $f, b$, height (Section \ref{sec:height}), and refine (Section \ref{sec:refine}).}}
\vspace{-0.8em}
\resizebox{\columnwidth}{!}{
\begin{tabular}{cccc|ccc|cc}
\toprule
\multicolumn{4}{c|}{\multirow{1}{*}{\textbf{$\text{LSTM}^{[\cdot]}$ components}}}  & \multicolumn{3}{c|}{\textbf{Synthetic}} & \multicolumn{2}{c}{\textbf{Real}} \\

$\varepsilon$ (EoT) & $f, b$ & height & refine  
& Mocap & Tennis & IPL & Mocap & IPL \\ \midrule

\xmark & \cmark & \cmark & \cmark
& 0.13 & 0.28 & 0.07 & 0.77 & 1.43 \\

\cmark & \xmark & \cmark & \cmark
& 0.06 & 0.17 & 0.11 & 0.84 & 2.23 \\

\cmark & \cmark & \xmark & \cmark
& 6.26 & 0.15 & 0.93 & 0.96 & 2.28 \\

\cmark & \cmark & \cmark & \xmark
& 0.10 & 0.18 & 0.053 & 1.13 & 1.84 \\

\xmark & \cmark & \cmark & \xmark
& 0.11 & 0.34 & 0.13 & 0.72 & 3.41 \\

\cmark & \xmark & \cmark & \xmark
& 0.09 & 0.17 & 1.09 & 0.87 & 3.13 \\

\cmark & \cmark & \xmark & \xmark
& 9.61 & 0.50 & 3.43 & 3.22 & 2.01 \\

\cmark & \cmark & \cmark & \cmark 
& \textbf{0.05} & \textbf{0.09} & \textbf{0.01} & \textbf{0.68} & \textbf{0.74} \\ \bottomrule
\end{tabular}}
\vspace{-2em}
\label{module_table}
\end{table}
\endgroup

\setlength{\tabcolsep}{4pt} 
\begin{table*}[t]
\begin{minipage}[b]{1.35\columnwidth}
\captionof{table}{
Comparison with prior work on the synthetic single-launch trajectory test set. We report NRMSEs $\pm$ S.E. for different levels of noise in the input 2D trajectory.
}
\vspace{-1em}
\centering
\resizebox{\textwidth}{!}{
\begin{tabular}[t]{l|c|c|c|c|c|c} 
    \toprule
    \multicolumn{1}{c|}{\multirow{3}{*}{\textbf{Method}}}         
    & \multicolumn{6}{c}{\textbf{Single-launch trajectory}} \\ 
    \multicolumn{1}{c|}{}                                                                         
    & No noise & $\pm5$ pixels & $\pm10$ pixels & $\pm15$ pixels & $\pm20$ pixels & $\pm25$ pixels \\ 
    \midrule
    Mocanu et al.\cite{mocanu2017estimating} & $1.02\pm{0.03}$ & $1.03\pm{0.04} $ & $1.05\pm{0.04}$ & $1.09\pm{0.38} $  & $1.15\pm{0.39}$ & $1.20\pm{0.42}$ \\
    Shen et al. \cite{shen20163d} & $0.11\pm{0.01}$ & $0.20\pm{0.01}$ & $0.30\pm{0.01}$ & $0.42\pm{0.01}$ & $0.53\pm{0.02}$ & $0.64\pm{0.02}$ \\
    \textbf{Ours} & \textbf{0.03 $\pm$ 0.002} & \textbf{0.05 $\pm$ 0.002} & \textbf{0.07$\pm$0.002} & \textbf{0.09$\pm$0.002} & \textbf{0.11$\pm$0.003} & \textbf{0.14$\pm$0.004} \\
    \bottomrule
  \end{tabular}}
    \vspace{-1em}
  \label{sota_table}
  \end{minipage}
  \hspace{0.1em}
\begin{minipage}[c]{0.75\columnwidth}
\centering
  \vspace{-1.7cm}

\includegraphics[width=0.95\textwidth]{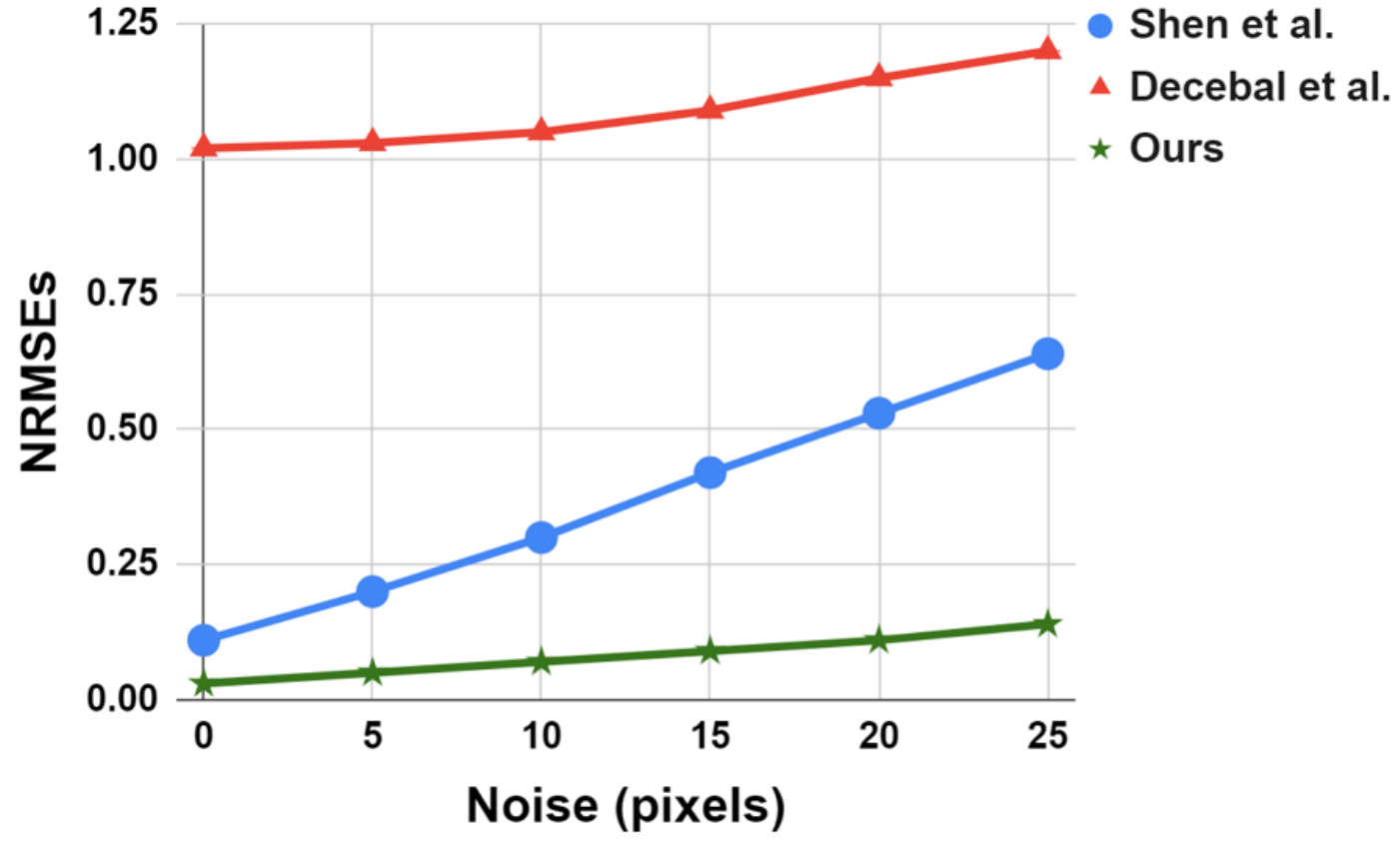}
\label{sota_noise_graph}
\end{minipage}
\vspace{-0.75cm}
\end{table*}

\vspace{-1em}
\subsubsection{Loss contributions}
We ablate each loss function and report the NRMSEs in Table \ref{loss_contrib_table} in Appendix. Removing $\mathcal{L}_{\varepsilon}$ and thus the EoT prediction altogether increases the NRMSEs by about 0.2 (8cm) on Synthetic tennis and 0.5 (80cm) on IPL dataset. Our simple constraint that enforces all the predictions to be above the ground $\mathcal{L}_{B}$ also helps improve accuracy across all datasets. Using $\mathcal{L}_\text{3D}$ alone performs the worst, while our full pipeline with all loss terms achieves the best performance.

\vspace{-1em}
    \subsubsection{Training / Fine-tuning on real data} We show the importance of leveraging synthetic data in Table \ref{syn_real_table} (Appendix). Training on Real Mocap data achieves an NRMSE of 0.29, compared to 0.17 when trained on Synthetic, both tested on the same real test set. Training on Synthetic then fine-tuning on Real achieves the best NRMSE of 0.08. This shows how simulation helps alleviate problems from small and noisy training data and can be useful for both scenarios where real data is or is not available.
\begingroup
\setlength{\tabcolsep}{5pt} 
\begin{table}[]
\centering
\small
\vspace{-1.1em}
\caption{\textbf{Comparison with SynthNet \cite{ertner2024synthnet} on TrackNet   \cite{huangtracknet}}}
\vspace{-0.9em}
\resizebox{\columnwidth}{!}{
\begin{tabular}{c|cc|cc|cc}
\toprule
\multirow{2}{*}{\centering TrackNet \cite{huangtracknet}} & \multicolumn{2}{c|}{T.acc (\%)} & \multicolumn{2}{c|}{T.F1} & \multicolumn{2}{c}{Landing Error (m.)} \\
& SynthNet & \textbf{Ours} & SynthNet  & \textbf{Ours} & SynthNet & \textbf{Ours} \\ \midrule
\multicolumn{1}{c|}{Game 1} & $64.15$ & \textbf{86.96} & $0.488$ & \textbf{0.873} & $3.19$ & \textbf{0.48} \\
\multicolumn{1}{c|}{Game 2} & $56.45$ & \textbf{96.72} & $0.402$ & \textbf{0.905} & $2.25$ & \textbf{0.53} \\
\multicolumn{1}{c|}{Game 3} & $54.54$ & \textbf{96.67} & $0.294$ & \textbf{0.952} & $3.59$ & \textbf{0.23} \\
\multicolumn{1}{c|}{Game 4} & $28.07$ & \textbf{73.91} & $0.187$ & \textbf{0.714} & $7.07$ & \textbf{0.96} \\
\multicolumn{1}{c|}{Game 5} & $45.45$ & \textbf{84.62} & $0.202$ & \textbf{0.841} & $3.30$ & \textbf{0.91} \\
\multicolumn{1}{c|}{Game 6} & $66.67$ & \textbf{89.74} & $0.388$ & \textbf{0.823} & $3.40$ & \textbf{0.73} \\
\multicolumn{1}{c|}{Game 7} & $54.00$ & \textbf{75.68} & $0.354$ & \textbf{0.547} & $2.91$ & \textbf{0.77} \\
\multicolumn{1}{c|}{Game 8} & $43.14$ & \textbf{85.42} & $0.320$ & \textbf{0.774} & $4.40$ & \textbf{0.90} \\
\multicolumn{1}{c|}{Game 9} & $59.52$ & \textbf{89.19} & $0.396$ & \textbf{0.803} & $2.61$ & \textbf{0.55} \\
\multicolumn{1}{c|}{Game 10} & $61.54$ & \textbf{93.18} & $0.552$ & \textbf{0.841} & $3.02$ & \textbf{0.28} \\
\midrule
\multicolumn{1}{c|}{Average} & $53.35$ & \textbf{87.21} & $0.358$ & \textbf{0.807} & $3.58$ & \textbf{0.63} \\
\bottomrule

\end{tabular}
}
\vspace{-1.5em}
\label{synthnet_tracknet}
\end{table}
\endgroup


  \vspace{-0.4em}
\section{Limitations \& Discussion}
Our method assumes the first and last frames are on the ground, which may require trimming the input sequence (e.g., starting after the first ground bounce following the serve). Despite this constraint, our method can handle any number of intermediate bounces or hits, as in  back-and-forth tennis rallies  (Figure~\ref{fig_tennis}). This contrasts with prior methods that require detecting all ground contact points to process each projectile separately. This assumption ensures the initial height is known (zero) during the height accumulation process. Rather than assuming it to be zero, predicting the initial height with a separate network is a promising direction for removing this manual step.

Our method may be affected by discrepancies between simulated and real distributions, particularly in unusual trajectories (Appendix~\ref{app:failurecases}). These stem partly from the simulation ignoring factors like spin, aerodynamics, and court type (e.g., grass), which affect friction and bounce. Incorporating these factors in future work could improve performance. Nonetheless, our learning-based approach remains more robust than competing methods and can handle more challenging scenarios beyond the single-launch trajectories typically tested in the literature.

In summary, we propose a method for 3D ball trajectory estimation from 2D monocular tracking. The key components of our learning-based pipeline are a novel 3D representation and intermediate representations that mitigate ambiguity in 3D prediction. These viewpoint-independent representations make the method well-suited for broadcast videos, where common camera angles are typically used, allowing us to train our network \emph{once} for reuse across multiple viewpoints. Extensive experiments show its effectiveness and generalization to challenging real-world scenarios, such as in sports games, despite training from simulation.
  \vspace{-0.5em}
\section*{Acknowledgement}
{
\spaceskip=0.3em plus 0.05em minus 0.2em
\xspaceskip=0.3em plus 0.05em
We thank Dr. Konstantinos Rematas for his valuable feedback, guidance, and assistance with revisions and figures. His work \cite{rematas2018soccer} and earlier explorations greatly inspired us and helped shape our approach.
}
  {
    \small
    \bibliographystyle{ieeenat_fullname}
    \bibliography{content/egbib.bib}
  }
    \clearpage
    
\newpage
\title{{\fontsize{13}{0}\selectfont\textbf{Appendix: Where Is The Ball: 3D Ball Trajectory Estimation From 2D Monocular Tracking}}}
\maketitle

\appendix 
\section{Overview}
\label{app:sup_mat}

In this Appendix, we present:
\begin{itemize}[noitemsep,topsep=0pt]
    \item Section \ref{app:simulation_details}: Simulation details.
    \item Section \ref{app:dataset_details}: Dataset details.
    \item Section \ref{app:net_arch}: Implementation details and network architectures.
    \item Section \ref{app:runtime}: Runtime.
    \item Section \ref{app:more_results}: Additional results.
    \item Section \ref{app:failurecases}: Failure cases.

\end{itemize}

\section{Simulation details}
\label{app:simulation_details}
We used Unity (v2019.3.2f1) with PhysX engine (v3.3) to simulate ball trajectories. The ground plane was created using a box collider object, and its center position was set to the origin. We used a sphere collider object with the property ``rigid'' for the ball. The camera parameters were manually set based on real-world parameters (estimated from three real datasets). For all simulations, we take into account the ball's size, weight, and a plausible range of ball speeds (by varying the applied force). Other factors, such as ball spin, aerodynamics, and court type (e.g., grass), are not considered in the current setup but could be incorporated in future work to enhance realism of the simulation, as they influence friction and bounce behavior.

\subsection{Mocap and IPL}
\begin{figure}[h]
\centering
  \includegraphics[scale=0.45]{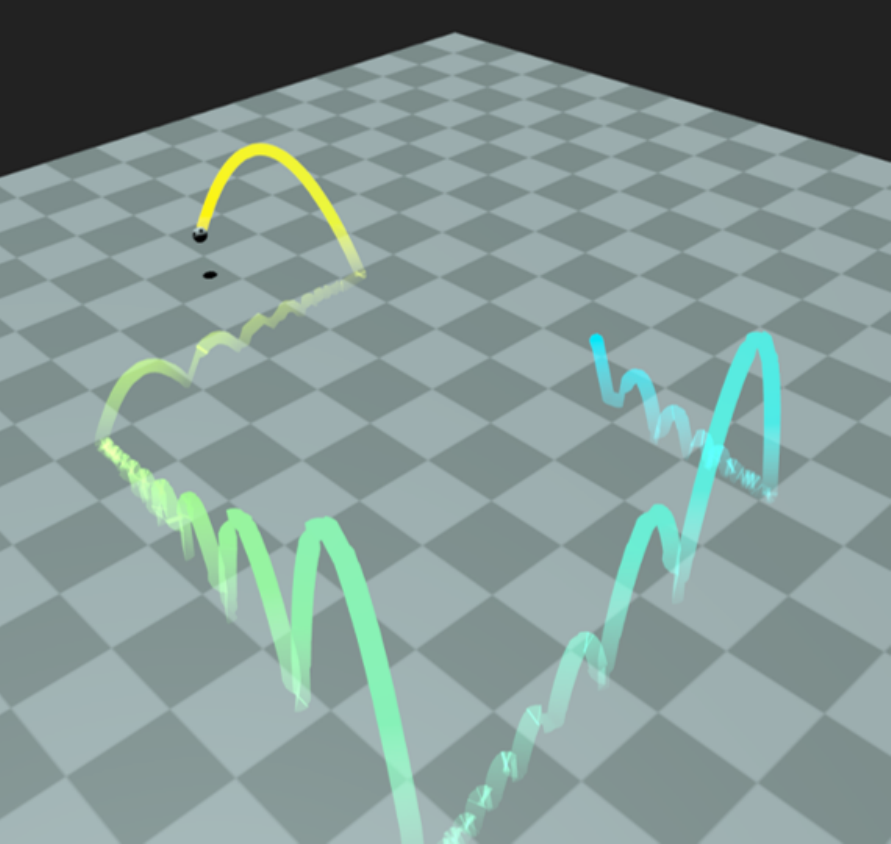}
  \caption{\textbf{Unity Game Engine for Mocap, IPL and Simpler synthetic datasets.}}
  \label{sim_mocap_ipl_simpler}
\end{figure}

To simulate a bouncing ball with multiple trajectories, we applied an impulse force at the beginning and let the ball bounce until its velocity dropped below a threshold, indicating that it had nearly stopped moving, before applying a new force. These forces had random magnitudes, and their directions were randomly generated so that projectile motions and rolling motions on the ground occurred with an equal chance. Note that projectile motions are generated using forces with positive y components, assuming y+ points upward, and rolling motions using forces with zero y component. The end-of-trajectory flag was only set true at the time step right before each force was being applied. The simulation of Mocap and IPL in Unity Game Engine is shown in Figure \ref{sim_mocap_ipl_simpler}.

\subsection{Tennis}
\begin{figure}[h]
\centering
  \includegraphics[scale=0.53]{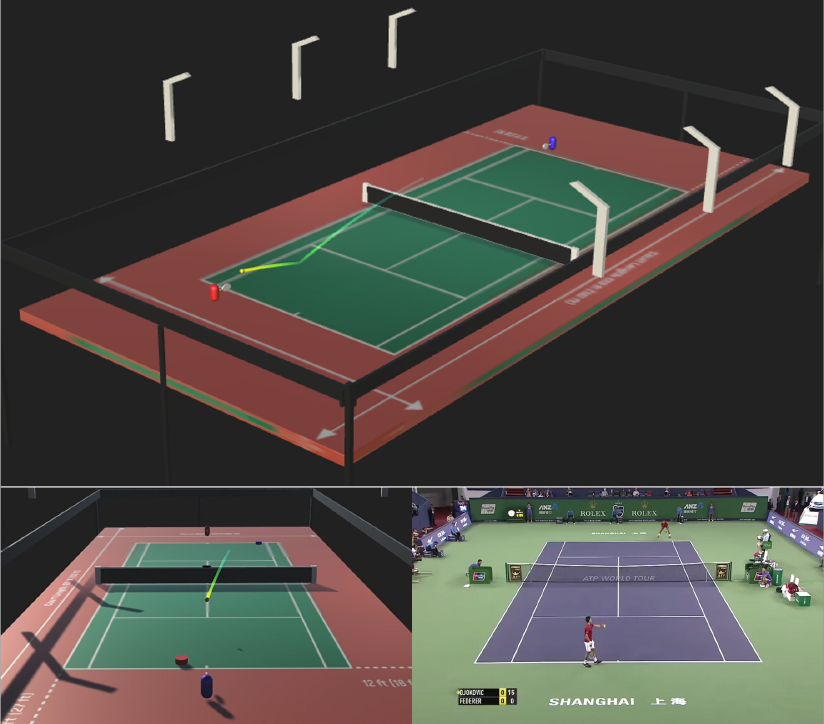}
  \caption{\textbf{Unity Game Engine for Tennis.}}
  \label{sim_tennis}
\end{figure}

To simulate tennis shots, we built upon an open-source tennis game \cite{tennis_game} and made the gameplay between two computer-bot players for the ease of data collection. Each bot has 2 actions: hit and receive. The hitter bot will randomly pick a location on the opponent's side for the ball to land and make a hit with random angles between 10-20 degrees (creating a lob shot or a flat shot). Then, the receiver bot will receive the ball behind the landing position with an offset of 5-7 meters away and subsequently becomes the hitter, and vice versa. The end-of-trajectory flag was only set true at the time step right before the bots make a hit. Additionally, the net was created using a box collider object for filtering out trajectories that are not passing over the net. The tennis simulation in Unity Game Engine is shown in Figure \ref{sim_tennis}.

\subsection{Synthetic Single-Launch Trajectory Dataset}\label{simplified_dataset}
For comparison with Mocanu et al. \cite{mocanu2017estimating} and Shen et al.\cite{shen20163d}, we matched their input assumptions (single force / single trajectory) and simulated single-trajectory sequences by launching the ball from the origin into a random direction in the first quadrant (0-90 degrees). Other configurations are similar to those used in Mocap and IPL.

\section{Dataset details}
\label{app:dataset_details}
In this section, we explain details of our synthetic datasets and real datasets. For synthetic datasets, the train, validation and test sets consist of 5000, 1500 and 500 sequences. 


\subsection{Synthetic Mocap} 
In this dataset, the sequence length varies between 85-2,873 time steps with an average of 460 time steps. The space for the ball to travel is about $11.11\times10.92\text{m}^2$ with the maximum height of 1.58m. Each sequence in the train set contains two consecutive trajectories, whereas in the validation and test sets, each sequence contains 1-7 consecutive trajectories.

\subsection{Synthetic IPL}
In this dataset, the sequence length varies between 37-949 time steps with an average of 104 time steps.
The space for the ball to travel is about $30\times75\text{m}^2$ with the maximum height of 10.42m. 
Similar to Synthetic Mocap, each sequence in the train set contains two consecutive trajectories, while in the validation and test sets, 1-7 consecutive trajectories.

\subsection{Synthetic Tracknet (Tennis)}
In this dataset, the sequence length varies between 64-822 time steps with an average of 122 time steps.
The space for the ball to travel is about $18.11\times37.12\text{m}^2$ with the maximum height of 3.87m. All sequences in train, validation, and test sets contain 3 strokes.

\subsection{Synthetic Single-Launch Trajectory Dataset}
\label{simplerdataset}
This synthetic dataset for state-of-the-art comparison with Mocanu et al. \cite{mocanu2017estimating} and Shen et al.\cite{shen20163d} has 300, 100 and 100 sequences for train, validation and test set. This dataset has the minimum and maximum sequence lengths of 46 and 106 time steps. The average sequence length of trajectories is 74 time steps. The space for the ball to travel is about $4.48\times4.43\text{m}^2$ with the maximum height of 0.77m. 

\subsection{Real IPL}
IPL soccer ball detection dataset \cite{fotouhi2017projection} contains short video streams of a real soccer match from 6 synchronized cameras at 25fps. The 2D ball tracking sequences are provided, and we followed the camera pose estimation pipeline in \cite{rematas2018soccer} to estimate the 3D ball positions used as ground truth. 
In this dataset, the 2D track points are sometimes missing for a few frames, e.g., due to occlusion. We describe our method to fill in these missing data in section \ref{app:auto_reg}. Then, we used the completed trajectories as input to our model.
There is a total of 9 remaining sequences that were successfully calibrated from the above process and satisfied our assumption that the ball starts and ends on the ground. The minimum and maximum sequence lengths are 18 and 147 and the average is 68 time steps. The ball travels within a space of size $30.60\times47.07\text{m}^2$ with the maximum height of $1.66$m.

\begin{figure}[h]
\centering
  \includegraphics[scale=0.7]{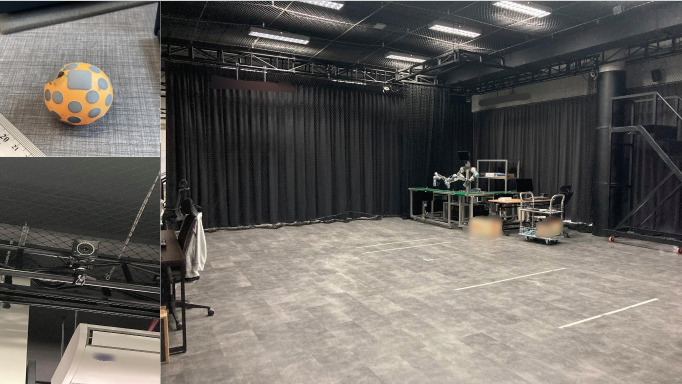}
  \caption{\textbf{Motion capture studio.} The top left is a ping-pong ball attached with IR reflective materials. The right image is our motion capture studio used to collect data. The bottom left is one of the eight IR cameras used in the studio.}
  \label{mocap_std}
\end{figure}

\subsection{Real Mocap}
This dataset captures the bouncing motion of a ping pong ball in a motion capture studio shown in Figure \ref{mocap_std}. This system uses 8 synchronized IR cameras to track IR reflective stickers that were attached on the ping-pong ball with a 40mm diameter. The camera frame rate was set to 50fps. The Mocap system provided the 3D positions of the ping-pong ball with a 2D tracking sequence from each camera with known parameters. We used all cameras with their 2D tracking sequences as input to our method for evaluation. We generated bouncing motions by throwing the ball upward within this space and kept re-throwing whenever the ball stopped moving from that last spot. This dataset contains 344 different trajectories, and the maximum number of consecutive trajectories is 3. The minimum and maximum sequence lengths are 150 and 907 and the average is 301 time steps.
The space for the ball to travel is about $6.21\times3.74\text{m}^2$ and the maximum height is 1.49m.

\subsection{Real Tracknet (Tennis)}
This dataset \cite{huangtracknet} contains 81 video clips of 10 tennis matches captured from a 30FPS broadcast camera. The 2D ball tracking annotations are also provided. We qualitatively evaluate our performance on 118 trajectories from 13 clips in one match. The minimum and maximum sequence lengths are 18 and 288, and the average is 92 time steps. Each sequence has a varying number of strokes between 1 to 10 (with an average of 3 strokes) and the tennis ball bounces 9 times at most (with an average of 4 bounces).

\section{Implementation details / Network architectures} \label{app:net_arch}

For training, we set $(\lambda_{\varepsilon}, \lambda_\text{3D}, \lambda_{B}) = (10, 1, 10)$ and trained our networks for 1,400 epochs using Adam optimizer~\cite{kingma2014adam} with a constant learning rate of 0.001 and a batch size of 256. 
We trained our LSTMs with backpropagation through time. Note that our trained pipeline can still predict output sequences of arbitrary lengths.
We also randomly add a Gaussian noise to each 2D input location ($u_t, v_t$) to simulate noisy 2D tracking from a tracking algorithm or human labels. Results for different levels of noise are reported in the main paper in Table \textcolor{red}{5}. 

Next, we explain the network architectures of:

\begin{enumerate}
\item EoT prediction network ($\text{LSTM}^{\varepsilon}$) in Table \ref{eot_arch}. 
\item Height prediction network ($\text{LSTM}^{\text{f, b}}$ and $\text{LSTM}^{\text{height}}$) in Table \ref{hagg_arch}, \ref{hbl_arch}.
\item Refinement network ($\text{LSTM}^\text{refine})$ in Table \ref{res_ref}.
\end{enumerate}
Note that in these tables, B is the batch size, L is the sequence length, and all LeakyReLUs use 0.01 slope.

\begingroup
\setlength{\tabcolsep}{13pt}
\begin{table}[h!]
\caption{Network architecture of the EoT prediction network ($\text{LSTM}^{\varepsilon}$).}
\centering
\resizebox{\columnwidth}{!}{%
\begin{tabular}{c|c|c}
\toprule
\textbf{Layer}    & \textbf{Activation} & \textbf{Output size}    \\ \midrule
Input    & -          & B x L x 4      \\ \hline
BiLSTM.0 & -          & B x L x 2 x 64 \\ \hline
\begin{tabular}[c]{@{}c@{}}BiLSTM.1\end{tabular}                            
& - & B x L x 2  x 64 \\ \hline
\begin{tabular}[c]{@{}c@{}} + output of BiLSTM.0 (residual) \\BiLSTM.2\end{tabular}               
& - & B x L x 2 x 64 \\ \hline
Concat   & -          & B x L x 128    \\ \hline
FC.0     & Leaky ReLU & B x L x 32     \\ \hline
FC.1     & Leaky ReLU & B x L x 32     \\ \hline
FC.2     & Leaky ReLU & B x L x 32     \\ \hline
FC.3     & Sigmoid    & B x L x 1      \\
\bottomrule
\end{tabular}}
\label{eot_arch}
\end{table}
\endgroup

\begingroup
\setlength{\tabcolsep}{10pt}
\begin{table}[h!]
\caption{Network architecture of the $\text{LSTM}^\text{f, b}$ in height prediction network. Note that we use the same architecture for both the forward and backward directions.}
\centering
\resizebox{5.5cm}{!}{%
\begin{tabular}{c|c|c}
\toprule
\textbf{Layer}    & \textbf{Activation} & \textbf{Output size}    \\ \midrule
Input    & -          & B x L x 6      \\ \hline
LSTM.0 & -          & B x L x 1 x 64 \\ \hline
\begin{tabular}[c]{@{}c@{}}LSTM.1\end{tabular}                            & - & B x L x 1 x 64 \\ \hline
\begin{tabular}[c]{@{}c@{}}LSTM.2\end{tabular}                            & - & B x L x 1 x 64 \\ \hline
Concat   & -          & B x L x 64    \\ \hline
FC.0     & Leaky ReLU & B x L x 32     \\ \hline
FC.1     & Leaky ReLU & B x L x 32     \\ \hline
FC.2     & Leaky ReLU & B x L x 32     \\ \hline
FC.3     & -          & B x L x 1      \\ 
\bottomrule
\end{tabular}}
\label{hagg_arch}
\vspace{-0.25em}
\end{table}
\endgroup

\begingroup
\setlength{\tabcolsep}{13pt}
\begin{table}[h!]
\caption{Network architecture of the $\text{LSTM}^\text{height}$ in height prediction network.}
\centering
\resizebox{\columnwidth}{!}{
\begin{tabular}{c|c|c}
\toprule
\textbf{Layer}    & \textbf{Activation} & \textbf{Output size}    \\ \midrule
Input    & -          & B x L x 5      \\ \hline
BiLSTM.0 & -          & B x L x 2 x 64 \\ \hline
\begin{tabular}[c]{@{}c@{}}BiLSTM.1\end{tabular}                            
& - & B x L x 2  x 64 \\ \hline
\begin{tabular}[c]{@{}c@{}} + output of BiLSTM.0 (residual) \\BiLSTM.2\end{tabular}               
& - & B x L x 2 x 64 \\ \hline
Concat   & -          & B x L x 128    \\ \hline
FC.0     & Leaky ReLU & B x L x 32     \\ \hline
FC.1     & Leaky ReLU & B x L x 32     \\ \hline
FC.2     & Leaky ReLU & B x L x 32     \\ \hline
FC.3     & -    & B x L x 1      \\ \bottomrule

\end{tabular}}
\vspace{-0.25em}
\label{hbl_arch}
\end{table}
\endgroup

\begingroup
\setlength{\tabcolsep}{13pt}
\begin{table}[h!]
\caption{Network architecture of the refinement network.}
\centering
\resizebox{\columnwidth}{!}{
\begin{tabular}{c|c|c}
\toprule
\textbf{Layer}    & \textbf{Activation} & \textbf{Output size}    \\ \midrule
Input    & -          & B x L x 7      \\ \hline
BiLSTM.0 & -          & B x L x 2 x 64 \\ \hline
\begin{tabular}[c]{@{}c@{}}BiLSTM.1\end{tabular}                            
& - & B x L x 2  x 64 \\ \hline
\begin{tabular}[c]{@{}c@{}} + output of BiLSTM.0 (residual) \\BiLSTM.2\end{tabular}               
& - & B x L x 2 x 64 \\ \hline
Concat   & -          & B x L x 128    \\ \hline
FC.0     & Leaky ReLU & B x L x 32     \\ \hline
FC.1     & Leaky ReLU & B x L x 32     \\ \hline
FC.2     & Leaky ReLU & B x L x 32     \\ \hline
FC.3     & -    & B x L x 3      \\ \bottomrule
\end{tabular}}
\vspace{-0.5cm}
\label{res_ref}
\end{table}
\endgroup

\subsection{Filling in missing track points (IPL dataset)}
\label{app:auto_reg}
In IPL dataset~\cite{fotouhi2017projection}, there are missing data points in some time steps in the 2D tracking sequences. We solve this problem with an additional pre-processing step that fills in the missing points before using the completed sequence as input to our main pipeline and other competing techniques. In particular, we trained an auto-regressive network also based on LSTMs that takes as input the temporal difference of 2D pixel coordinates $(\Delta{u_t}, \Delta{v_t})$ and predicts the difference for the next time step $(\Delta{u_{t+1}},\Delta{v_{t+1}})$, following \cite{graves2013generating}. This network consists of 2 independent directional-LSTMs that auto-regress the sequence in the forward and backward directions shown in Table \ref{ar_arch}. The resulting two predicted sequences are combined with linear ramp weighting similar to Eq. 3 in the main paper, to output the final 2D tracking sequence. Note that if a tracking data point is available for the current time step, we simply use it. We trained this network with the teacher forcing technique \cite{williams1989learning}.

\begingroup
\setlength{\tabcolsep}{13pt}
\begin{table}[h!]
\caption{Network architecture of the auto-regressive model for interpolating missing data points. Note that we used the same architecture for both forward and backward directions.}
\centering
\resizebox{\columnwidth}{!}{
\begin{tabular}{c|c|c}
\toprule
\textbf{Layer}    & \textbf{Activation} & \textbf{Output size}    \\ \midrule
Input    & -          & B x L x 2      \\ \hline
LSTM.0 & -          & B x L x 64 \\ \hline
\begin{tabular}[c]{@{}c@{}}LSTM.1\end{tabular}                            & - & B x L x 64 \\ \hline
\begin{tabular}[c]{@{}c@{}} + output of LSTM.0 (residual) \\ LSTM.2 \end{tabular}               & - & B x L x 64 \\ \hline
\begin{tabular}[c]{@{}c@{}} + output of LSTM.0 and LSTM.1 (residual) \\ 
LSTM.3 \end{tabular} & - & B x L x 64 \\ \hline
FC.0     & Leaky ReLU & B x L x 64     \\ \hline
FC.1     & Leaky ReLU & B x L x 32     \\ \hline
FC.2     & Leaky ReLU & B x L x 16     \\ \hline
FC.3     & Leaky ReLU & B x L x 8      \\ \hline
FC.4     & Leaky ReLU & B x L x 4      \\ \hline
FC.5     & -          & B x L x 2      \\ \bottomrule
\end{tabular}}
\label{ar_arch}
\end{table}
\endgroup

\section{Runtime}
\label{app:runtime}
We measured runtime on the test set of Simpler Synthetic dataset (Appendix \ref{simplerdataset}), which contains 100 trajectories (7,463 timesteps in total). We tested our method and other competing techniques on 100 trajectories for 100 times (10,000 sequences) on a desktop with AMD Ryzen 9 3900X and a single NVIDIA 2080 super. Our method took an average of $1.01\pm{0.11}$ms per frame, which is about $8.6\times$ faster than the other learning-based Mocanu et al. \cite{mocanu2017estimating} ($8.7\pm{1}$ms). The physics-based method, Shen et al. \cite{shen20163d}, only requires optimization and is the fastest with an average runtime of $0.012\pm{0.003}$ms per frame.

\section{Additional results}
\label{app:more_results}
In this section, we provide an additional prior work comparison on two real datasets, additional error metrics, as well as additional qualitative results for three real and three synthetic datasets.
\subsection{Comparison with prior work on Real Mocap and Real IPL}
We compare our method to the same state-of-the-art methods \cite{mocanu2017estimating, shen20163d} used in Section 4.2 of the main paper, but each test example in this experiment contains multiple trajectories due to multiple acting forces (e.g., tennis hits).
Note again that these prior methods are not designed for multiple trajectories, but we include this experiment for completeness. We performed a fair comparison using a single-launch trajectory test set in Section 4.2.

Table \ref{sota_comp_real} reports distance NRMSEs on the test sets of Real Mocap and IPL datasets. 
Our method achieves significantly better NRMSEs with performance gaps of up to 75.4 in Mocap and 13.6 in IPL, but this is expected as these test scenarios violate their assumptions.

\subsection{Results using other NRMSE variants}

Table \ref{nrmse_wrt_table} reports different variants of NRMSEs, which are RMSEs $\times100\%$ divided by the trajectory height, area's length, area's width, or the distance to camera, following \cite{calandre2021extraction}. Here the length and width are the field dimensions (e.g., tennis court ($23.27\times10.97 m^2$) or soccer pitch ($105\times69.5 m^2$). We report NRMSEs for \emph{distance}, based on the L2 distance on the xyz coordinates, and \emph{height}, based on the distance along the y coordinate only. Since our method may exhibit errors relative to the size of the playing area, these metrics are important for assessing our performance for different applications or different world scales. For example, when visualizing the soccer ball in the entire soccer field, errors with respect to the area's length or width may be appropriate.

For Real Mocap, we achieve a 0.48\% distance NRMSE with respect to both the area's length and width. For IPL, the errors are 1.13\% and 1.71\% with respect to the soccer pitch's dimensions, or 1.13\% with respect to the camera distance, which is about 106m away from the soccer pitch.

\subsection{Other quantitative metrics}
We show quantitative results from all experiments and ablation studies in RMSEs (in centimeters) in Table \ref{features_table_rmse}-\ref{sota_table_rmse}. Additionally, we report the statistics of ground penetration in the predicted trajectories on Real Tracknet in Table \ref{tracknet_qual_stat}.

\begin{table*}[ht]
\centering
\setlength{\tabcolsep}{2.5pt} 
\begin{minipage}[t]{0.48\textwidth}
\centering
\caption{\textbf{Comparison with prior work on Real Mocap and Real IPL.} The numbers are NRMSEs. Note that each test example in these datasets contains multiple trajectories, which are outside prior work's assumptions.}
\vspace{-0.5em}
\resizebox{0.5\linewidth}{!}{
\begin{tabular}{l|c|c}
\toprule
\multicolumn{1}{c|}{\multirow{2}{*}{Method}} & \multicolumn{2}{c}{\textbf{Real}} \\
& Mocap & IPL \\ \midrule
Mocanu et al.\cite{mocanu2017estimating} & 15.92 & 14.33 \\
Shen et al.\cite{shen20163d} & 76.09 & 5.03 \\
\textbf{Ours} & \textbf{0.68} & \textbf{0.74} \\
\bottomrule
\end{tabular}}
\label{sota_comp_real}
\end{minipage}
\hfill
\begin{minipage}[t]{0.48\textwidth}
\centering
\setlength{\tabcolsep}{7pt} 
\caption{\textbf{How helpful is simulated data?} 
We report NRMSEs $\pm$ S.E. of training on Real and Synthetic Mocap, as well as on Synthetic then fine-tuning on Real. Using Synthetic for training or pre-training outperforms training on Real alone.}
\vspace{-0.5em}
\resizebox{0.95\linewidth}{!}{
\begin{tabular}{c|cc}
\toprule
\textbf{Training data} & \textbf{Distance} & \textbf{Height} \\ \midrule
Real Mocap & $0.29 \pm{0.04}$ & $7.99 \pm{1.68}$ \\
Synthetic Mocap & $0.17 \pm{0.01}$ & $7.12 \pm{1.14}$ \\
Synthetic + Real (Fine-tuned) & \textbf{$0.08 \pm{0.004}$} & \textbf{$5.23 \pm{1.06}$} \\
\bottomrule
\end{tabular}}
\label{syn_real_table}
\end{minipage}
\vspace{-1em}
\end{table*}

\begingroup
\setlength{\tabcolsep}{2pt}
\begin{table*}[t!]
\centering
\caption{We report NRMSEs with respect to the trajectory height, area's length, area's width, and distance to camera, following Calendre et al. \cite{calandre2021extraction}. *Each row shaded in gray shows the denominators (meter) used to compute each normalized RMSE.}
\resizebox{0.87\textwidth}{!}{
\begin{tabular}{l|cc|cc|cc|cc|cc}
\toprule
\multicolumn{1}{c|}{\multirow{3}{*}{\textbf{Metric}}}         
& \multicolumn{6}{c|}{\textbf{Synthetic}} & \multicolumn{4}{c}{\textbf{Real}} \\ 
\multicolumn{1}{c|}{}                                                                         
& \multicolumn{2}{c}{Mocap} & \multicolumn{2}{c}{Tennis} & \multicolumn{2}{c|}{IPL}            
& \multicolumn{2}{c}{Mocap} & \multicolumn{2}{c}{IPL}   \\ \cmidrule{2-11}
\multicolumn{1}{c|}{}                                                                         & \multicolumn{1}{c}{Distance} & \multicolumn{1}{c|}{Height} & \multicolumn{1}{c}{Distance} & \multicolumn{1}{c|}{Height} & \multicolumn{1}{c}{Distance} & \multicolumn{1}{c|}{Height} & \multicolumn{1}{c}{Distance} & \multicolumn{1}{c|}{Height} & \multicolumn{1}{c}{Distance} & \multicolumn{1}{c}{Height} \\ \midrule

RMSE (cm) & 1.33 & 0.48 & 8.48 & 2.25 & 3.43 & 0.80 & 3.83 & 2.15 & 119.15 & 26.04 \\
\midrule
\rowcolor{Gray!40}
Trajectory height (m) & \multicolumn{2}{c|}{1.58} & \multicolumn{2}{c|}{3.87} & \multicolumn{2}{c|}{10.42} &  \multicolumn{2}{c|}{1.49} & \multicolumn{2}{c}{1.66} \\
\%NRMSE  & 0.84 & 0.30 & 2.19 &	0.58 & 0.33 & 0.08 & 2.59 & 1.45 & 71.77 & 15.68 \\ \midrule

\rowcolor{Gray!40}
Area's length (m) & \multicolumn{2}{c|}{10.92} & \multicolumn{2}{c|}{23.77} & \multicolumn{2}{c|}{75.00} & \multicolumn{2}{c|}{8.00} & \multicolumn{2}{c}{105.00} \\
\%NRMSE & 0.12 & 0.04 & 0.36 & 0.09 & 0.05 & 0.01 & 0.48 & 0.27 & 1.13 & 0.25 \\
\midrule

\rowcolor{Gray!40}
Area's width (m) & \multicolumn{2}{c|}{11.11} & \multicolumn{2}{c|}{10.97} & \multicolumn{2}{c|}{30.00} & \multicolumn{2}{c|}{8.00} & \multicolumn{2}{c}{69.50} \\
\%NRMSE & 0.12 & 0.04 & 0.77 & 0.21 & 0.11 & 0.03 & 0.48 & 0.27 & 1.71 & 0.37 \\
\midrule

\rowcolor{Gray!40}
Distance to camera (m) & \multicolumn{2}{c|}{6.37} & \multicolumn{2}{c|}{32.84} & \multicolumn{2}{c|}{105.66} & \multicolumn{2}{c|}{6.37} & \multicolumn{2}{c}{105.66}\\
\%NRMSE & 0.21 & 0.08 &	0.26 & 0.07 & 0.03 & 0.01 & 0.60 & 0.34 & 1.13 & 0.25 \\
\midrule

\%NRMSE (RMSE / (max - min)) & 0.09 & - & 0.15 & - & 0.02 & - & 1.01 & - & 1.03 & - 
\\\bottomrule
\end{tabular}}
\label{nrmse_wrt_table}
\vspace{1cm}
\end{table*}
\endgroup

\begingroup
\setlength{\tabcolsep}{1.5pt}
\begin{table*}[t!]
\caption{\textbf{Ablation study on input/output parameterization}. We evaluate our full pipeline with different types of input / output parameterization. The numbers are RMSEs of distance and height measured in centimeter.}
\centering
\resizebox{\textwidth}{!}{
\begin{tabular}{cc|cc|cc|cc|cc|cc}
\toprule
\multicolumn{2}{c|}{\multirow{2}{*}{Parameterization}}         & \multicolumn{6}{c|}{\textbf{Synthetic}} & \multicolumn{4}{c}{\textbf{Real}} \\
\multicolumn{2}{c|}{}                                                                         & \multicolumn{2}{c}{Mocap}                                 & \multicolumn{2}{c}{Tennis}                                & \multicolumn{2}{c|}{IPL}                                   & \multicolumn{2}{c}{Mocap}                                 & \multicolumn{2}{c}{IPL}                                   \\ \cmidrule{3-12}

\multicolumn{1}{c}{Input} & \multicolumn{1}{c|}{Output (before refine.)} &          
\multicolumn{1}{c}{Distance} & \multicolumn{1}{c|}{Height} & \multicolumn{1}{c}{Distance} & \multicolumn{1}{c|}{Height} & \multicolumn{1}{c}{Distance} & \multicolumn{1}{c|}{Height} & \multicolumn{1}{c}{Distance} & \multicolumn{1}{c|}{Height} & \multicolumn{1}{c}{Distance} & \multicolumn{1}{c}{Height} \\ \cmidrule{1-12}

\multirow{2}{*}{Pixel} &  xyz 
& 165.19 & 9.49 & 18.43 & 5.12 & 117.25 & 27.44 & 91.63 & 28.21 & 240.81 & 44.88 \\ 
{}& height
& 2.82 & 0.72 & 19.29 & 3.96 & 13.78 & 2.98 & 52.12 & 24.77 & 265.79 & 50.62 \\ \cline{1-12}

\multirow{2}{*}{Pixel + $E$} & xyz
& 96.13 & 75.14 & 34.96 & 11.57 & 81.39 & 17.63 & 160.22 & 28.02 & 221.26 & 39.41 \\  

{} & height 
& 3.54 & 1.11 & 16.78 & 3.46 & 8.25 & 1.95 & 52.36 & 25.07 & 273.175 & 53.17 \\ \cline{1-12}

\multirow{2}{*}{$\mathbf{p}_\text{ground}$ + $(\varphi_{az}, \theta_{el})$} & xyz 
& 8.33 & 6.03 & 13.52 & 3.22 & 73.7 & 16.06 & 51.95 & 11.92 & 225.50 & 44.65 \\  

{} & height
& 4.11 & 1.29 & 13.51 & 2.68 & 4.53 & 1.21 & 28.90 & 13.67 & 248.31 & 47.21 \\  \cline{1-12}

\multirow{2}{*}{$\mathbf{p}_\text{ground}$ + $(\varphi_{az}^{\sin{},\cos{}}, \theta_{el}^{\sin{},\cos{}})$} & xyz
& 2.25 & 1.18 & 14.06 & 3.45 & 127.58 & 29.86 & 21.89 & 6.04 & 271.52 & 55.25 \\  

{} & height
& 13.48 & 5.47 & 12.33 & 2.45 & 5.78 & 1.46 & 6.77 & 3.50 & 130.27 & 27.28 \\  \cline{1-12}

\multirow{2}{*}{$\mathbf{p}_\text{ground}$ + $\mathbf{p}_\text{vertical}$ \textbf{(Ours)}} & xyz
& 2.23 & 1.12 & 13.02 & 3.12 & 83.02 & 22.59 & 5.13 & 2.57 & 296.55 & 56.96 \\  

{} & height \textbf{(Ours)}

& \textbf{1.33} & \textbf{0.48} & \textbf{8.48} & \textbf{2.25} & \textbf{3.43} & \textbf{0.80} & \textbf{3.83} & \textbf{2.15} & \textbf{119.15} & \textbf{26.04} \\  \bottomrule

\end{tabular}
}
\label{features_table_rmse}
\vspace{1cm}
\end{table*}
\endgroup

\begingroup
\setlength{\tabcolsep}{1.5pt}
\begin{table*}[t!]
\caption{\textbf{Ablation study on LSTM components}. The numbers are RMSEs of distance and height measured in centimeter.}
\centering
\resizebox{0.8\textwidth}{!}{
\begin{tabular}{cccc|cc|cc|cc|cc|cc}
\toprule
\multicolumn{4}{c|}{\multirow{2}{*}{$\text{LSTM}^{[*]}$ components}}  & \multicolumn{6}{c|}{\textbf{Synthetic}} & \multicolumn{4}{c}{\textbf{Real}} \\ 
\multicolumn{4}{c|}{} & \multicolumn{2}{c}{Mocap} & \multicolumn{2}{c}{Tennis} & \multicolumn{2}{c|}{IPL} & \multicolumn{2}{c}{Mocap} & \multicolumn{2}{c}{IPL}  \\ \cmidrule{5-14}
$\varepsilon$ (EoT) & $f, b$ & height & refine                                     & \multicolumn{1}{c}{Distance} & \multicolumn{1}{c|}{Height} & \multicolumn{1}{c}{Distance} & \multicolumn{1}{c|}{Height} & \multicolumn{1}{c}{Distance} & \multicolumn{1}{c|}{Height} & \multicolumn{1}{c}{Distance} & \multicolumn{1}{c|}{Height} & \multicolumn{1}{c}{Distance} & \multicolumn{1}{c}{Height} \\ \midrule
\xmark & \cmark & \cmark & \cmark
& 19.40 & 0.93 & 18.53 & 3.63 & 3.75 & 0.95 & 6.16 & 3.16 & 228.28 & 44.52\\

\cmark & \xmark & \cmark & \cmark
& 3.58 & 1.05 & 13.33 & 2.61 & 4.30 & 1.04 & 4.10 & 3.35 & 290.21 & 56.01 \\

\cmark & \cmark & \xmark & \cmark
& 53.77 & 30.89 & 11.92 & 2.46 & 182.01 & 46.23 & 6.83 & 3.11 & 284.33 & 56.22 \\

\cmark & \cmark & \cmark & \xmark
& 2.80 & 0.92 & 11.89 & 2.37 & 8.66 & 1.93 & 4.71 & 3.86 & 272.84 & 54.21\\

\xmark & \cmark & \cmark & \xmark
& 4.96 & 1.91 & 20.73 & 4.17 & 14.54 & 3.39 & 4.92 & 2.63 & 374.85 & 73.09\\

\cmark & \xmark & \cmark & \xmark
& 3.57 & 1.04 & 14.07 & 2.85 & 47.87 & 2.08 & 6.68 & 3.58 & 315.66 & 60.85 \\

\cmark & \cmark & \xmark & \xmark
& 125.4 & 53.33 & 32.01 & 7.07 & 555.15 & 128.91 & 29.79 & 16.17 & 258.76 & 52.13 \\

\cmark & \cmark & \cmark & \cmark
& \textbf{1.33} & \textbf{0.48} & \textbf{8.48} & \textbf{2.25} & \textbf{3.43} & \textbf{0.80} & \textbf{3.83} & \textbf{2.15} & \textbf{119.15} & \textbf{26.04}\\ 
\bottomrule
\end{tabular}}
\label{module_table_rmse}
\vspace{0.5cm}
\end{table*}
\endgroup

\begingroup
\setlength{\tabcolsep}{7pt} 
\begin{table*}[ht]
\centering
\caption{\textbf{Ablation study on loss terms}. We train our full pipeline with each loss term removed and report distance NRMSEs.}
\vspace{-0.8em}
\resizebox{1\columnwidth}{!}{
\begin{tabular}{l|c|c|c|c|c}
\toprule
\multicolumn{1}{c|}{\multirow{2}{*}{Loss}} & \multicolumn{3}{c|}{\textbf{Synthetic}} & \multicolumn{2}{c}{\textbf{Real}} \\
& \multicolumn{1}{c}{Mocap}  & \multicolumn{1}{c}{Tennis} & \multicolumn{1}{c|}{IPL}
& \multicolumn{1}{c}{Mocap}  & \multicolumn{1}{c}{IPL} \\ \midrule
no $\mathcal{L}_{\varepsilon}$  
& 0.15 & 0.25 & 0.06 & 0.84 & 1.24 \\
no $\mathcal{L}_{B}$ 
& 0.09 & 0.27 & 0.05 & 0.87 & 1.34 \\
no $\mathcal{L}_{\varepsilon}$, $\mathcal{L}_{B}$ 
& 0.23 & 0.29 & 0.08 & 0.98 & 3.16 \\
\textbf{Ours} (all terms)
& \textbf{0.05} & \textbf{0.09} & \textbf{0.01} & \textbf{0.68} & \textbf{0.74} \\ \bottomrule
\end{tabular}}
\label{loss_contrib_table}
\end{table*}
\endgroup

\begingroup
\setlength{\tabcolsep}{1.5pt}
\begin{table*}[h!]
\caption{\textbf{Ablation study on loss terms}. We train our full pipeline with each loss term removed. The numbers are RMSEs of distance and height measured in centimeter.}
\centering
\resizebox{0.75\textwidth}{!}{
\begin{tabular}{l|cc|cc|cc|cc|cc}
\toprule
\multicolumn{1}{c|}{\multirow{3}{*}{Loss}}         & \multicolumn{6}{c|}{\textbf{Synthetic}} & \multicolumn{4}{c}{\textbf{Real}} \\ 
\multicolumn{1}{c|}{}                                                                         & \multicolumn{2}{c}{Mocap} & \multicolumn{2}{c}{Tennis} & \multicolumn{2}{c|}{IPL}            
& \multicolumn{2}{c}{Mocap} & \multicolumn{2}{c}{IPL}   \\ \cmidrule{2-11}
\multicolumn{1}{c|}{}                                                                         & \multicolumn{1}{c}{Distance} & \multicolumn{1}{c|}{Height} & \multicolumn{1}{c}{Distance} & \multicolumn{1}{c|}{Height} & \multicolumn{1}{c}{Distance} & \multicolumn{1}{c|}{Height} & \multicolumn{1}{c}{Distance} & \multicolumn{1}{c|}{Height} & \multicolumn{1}{c}{Distance} & \multicolumn{1}{c}{Height} \\ \midrule

no $\mathcal{L}_{\varepsilon}$  & 7.05 & 4.86 & 16.39 & 3.23 & 7.80 & 1.87 & 4.02 & 2.26 & 179.98 & 36.21 \\
no $\mathcal{L}_{B}$ & 5.74 & 4.64 & 16.73 & 3.36 & 8.94 & 2.34 & 4.16 & 2.65 & 214.28 & 41.86 \\
no $\mathcal{L}_{\varepsilon}$, $\mathcal{L}_{B}$ & 9.03 & 3.4 & 17.46 & 3.5 & 15.61 & 3.38 & 4.2 & 2.45 & 439.63 & 86.29 \\
\textbf{Ours} (all terms) &
  \textbf{1.33} &
  \textbf{0.48} &
  \textbf{8.48} &
  \textbf{2.25} &
  \textbf{3.43} &
  \textbf{0.80} &
  \textbf{3.83} &
  \textbf{2.15} & 
  \textbf{119.15} & 
  \textbf{26.04}
  \\ \bottomrule
\end{tabular}
}
\label{loss_contrib_table_rmse}

\end{table*}
\endgroup



\begingroup
\setlength{\tabcolsep}{1.5pt}
\begin{table*}[h!]
\centering
\caption{\textbf{Comparison with prior work on Synthetic Mocap}. The numbers are RMSEs of distance and height measured in centimeter for varying levels of noise in the input 2D trajectory. 
}
\resizebox{0.92\textwidth}{!}{
\begin{tabular}{l|cc|cc|cc|cc|cc|cc}
\toprule
\multicolumn{1}{c|}{\multirow{3}{*}{Method}}         
& \multicolumn{12}{c}{\textbf{Synthetic Mocap}} \\ 
\multicolumn{1}{c|}{}                                                           
& \multicolumn{2}{c|}{No noise} & \multicolumn{2}{c|}{$\pm{5}$ pixels} & \multicolumn{2}{c|}{$\pm{10}$ pixels}            
& \multicolumn{2}{c|}{$\pm{15}$ pixels} & \multicolumn{2}{c|}{$\pm{20}$ pixels} &\multicolumn{2}{c}{$\pm{25}$ pixels} \\ \cmidrule{2-13}
\multicolumn{1}{c|}{}                                                                         & \multicolumn{1}{c}{Distance} & \multicolumn{1}{c|}{Height} & \multicolumn{1}{c}{Distance} & \multicolumn{1}{c|}{Height} & \multicolumn{1}{c}{Distance} & \multicolumn{1}{c|}{Height} & \multicolumn{1}{c}{Distance} & \multicolumn{1}{c|}{Height} & \multicolumn{1}{c}{Distance} & \multicolumn{1}{c|}{Height} & \multicolumn{1}{c}{Distance} & \multicolumn{1}{c}{Height} \\ \midrule

Mocanu et al.\cite{mocanu2017estimating} & 8.58	& 6.44 & 8.62 & 6.45 & 8.72	& 6.50 & 8.91 & 7.11 & 9.21 & 7.45 & 9.41 & 7.72\\
Shen et al. \cite{shen20163d} & 1.83 & 1.37 & 2.10 & 1.50 & 2.80 & 1.83 & 3.75 & 2.34 & 4.86 & 2.96 & 5.65 & 3.37\\
\textbf{Ours} & \textbf{0.60} & \textbf{0.30} & \textbf{0.65} & \textbf{0.32} & \textbf{0.69} & \textbf{0.34} & \textbf{0.97} & \textbf{0.36} & \textbf{1.30} & \textbf{0.38} & \textbf{1.66} & \textbf{0.45} \\\bottomrule
\end{tabular}
}
\label{sota_table_rmse}
\end{table*}
\endgroup

\begingroup
\setlength{\tabcolsep}{2pt}
\begin{table*}[t]
\centering
\caption{\textbf{Qualitative analysis on Real Tracknet (Tennis).} We report the statistics of points mistakenly predicted below ground at different penetration distances. 
}
\resizebox{0.93 \textwidth}{!}{
\begin{tabular}{l|c|c|c|c|c|c}
\toprule
\multicolumn{1}{c|}{\multirow{2}{*}{Metric}}         
& \multicolumn{6}{c}{\textbf{Real Tracknet (Tennis)}} \\ 
\multicolumn{1}{c|}{}                                                           
& \multicolumn{1}{c|}{0-2.5 cm.} & \multicolumn{1}{c|}{2.5-5 cm.} & \multicolumn{1}{c|}{5 - 7.5 cm.}            
& \multicolumn{1}{c|}{7.5 - 10 cm.} & \multicolumn{1}{c|}{10 - 25 cm.} &\multicolumn{1}{c}{25 - 50 cm.} \\ \cmidrule{1-7}

\#(predicted points below ground) {\color{gray}(N=181)} & 49 & 51 & 29 & 17 & 34 & 1 \\ \cmidrule{1-7}
\ \ \ \ as a percentage of \#(ground contact points) {\color{gray}(N=236)} & 20.76\% & 21.61\% & 12.29\% & 7.20\% & 14.41\% & 0.42\% \\
\ \ \ \ as a percentage of \#(all points) {\color{gray}(N=10,844)} & 0.45\% & 0.47\% & 0.27\% & 0.16\% & 0.31\% & 0.01\% \\
\bottomrule
\end{tabular}
}
\label{tracknet_qual_stat}
\end{table*}
\endgroup

\subsection{Qualitative results}

\begin{figure*}[h]
\centering
  \includegraphics[width=\textwidth]{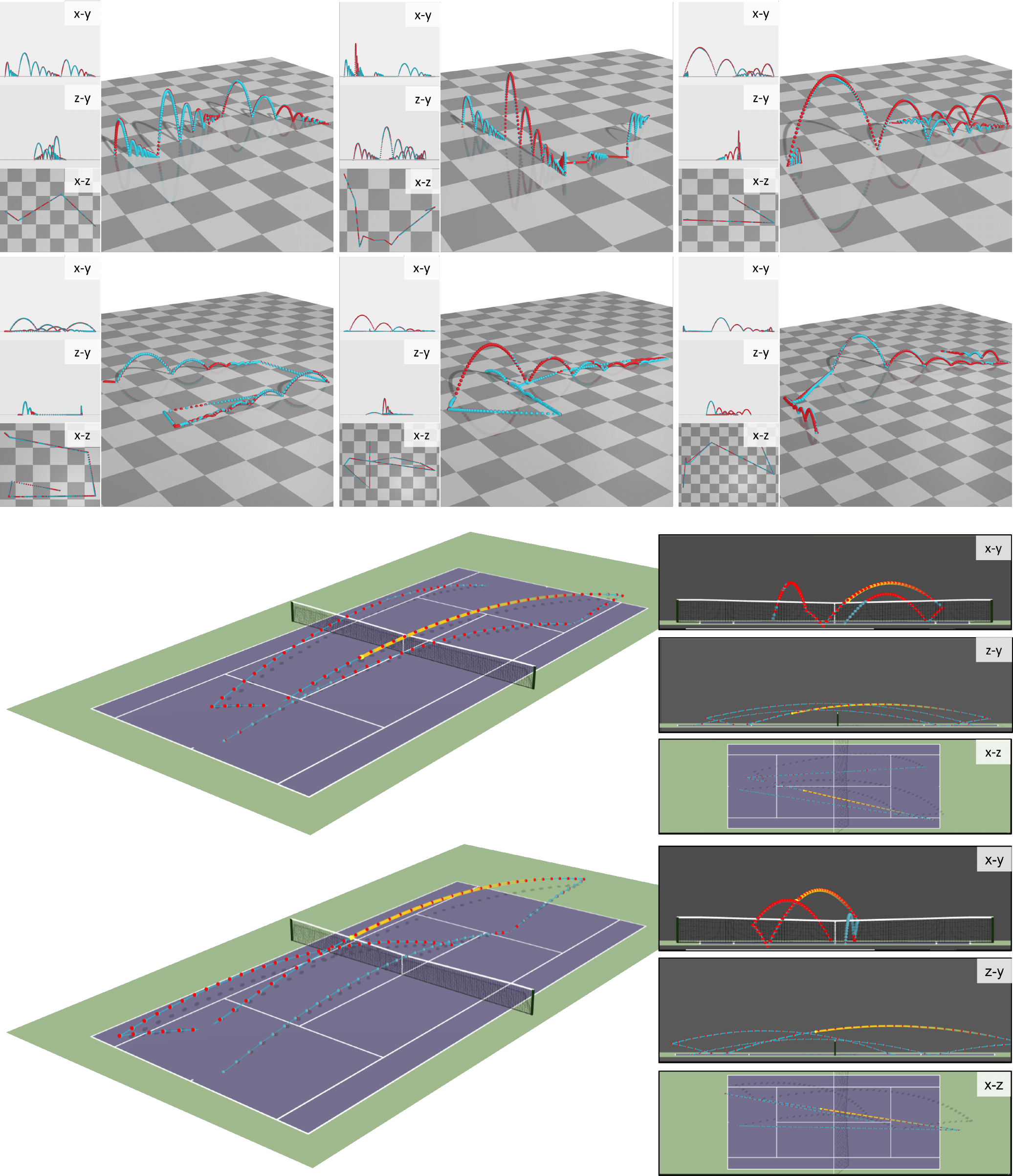}
  \caption{\textbf{Qualitative results on synthetic datasets.} Blue: our predictions. Red: ground truth.
  The first row is the results from Synthetic Mocap and each checkerboard block is 75$\times$75 cm$^2$. The second row is the results from Synthetic IPL and each checkerboard block is 250$\times$250 cm$^2$. The last two rows are the results from Synthetic Tennis.}
  \label{sm_synthetic_all}
\end{figure*}

\begin{figure*}[h]
\centering
  \includegraphics[width=\textwidth]{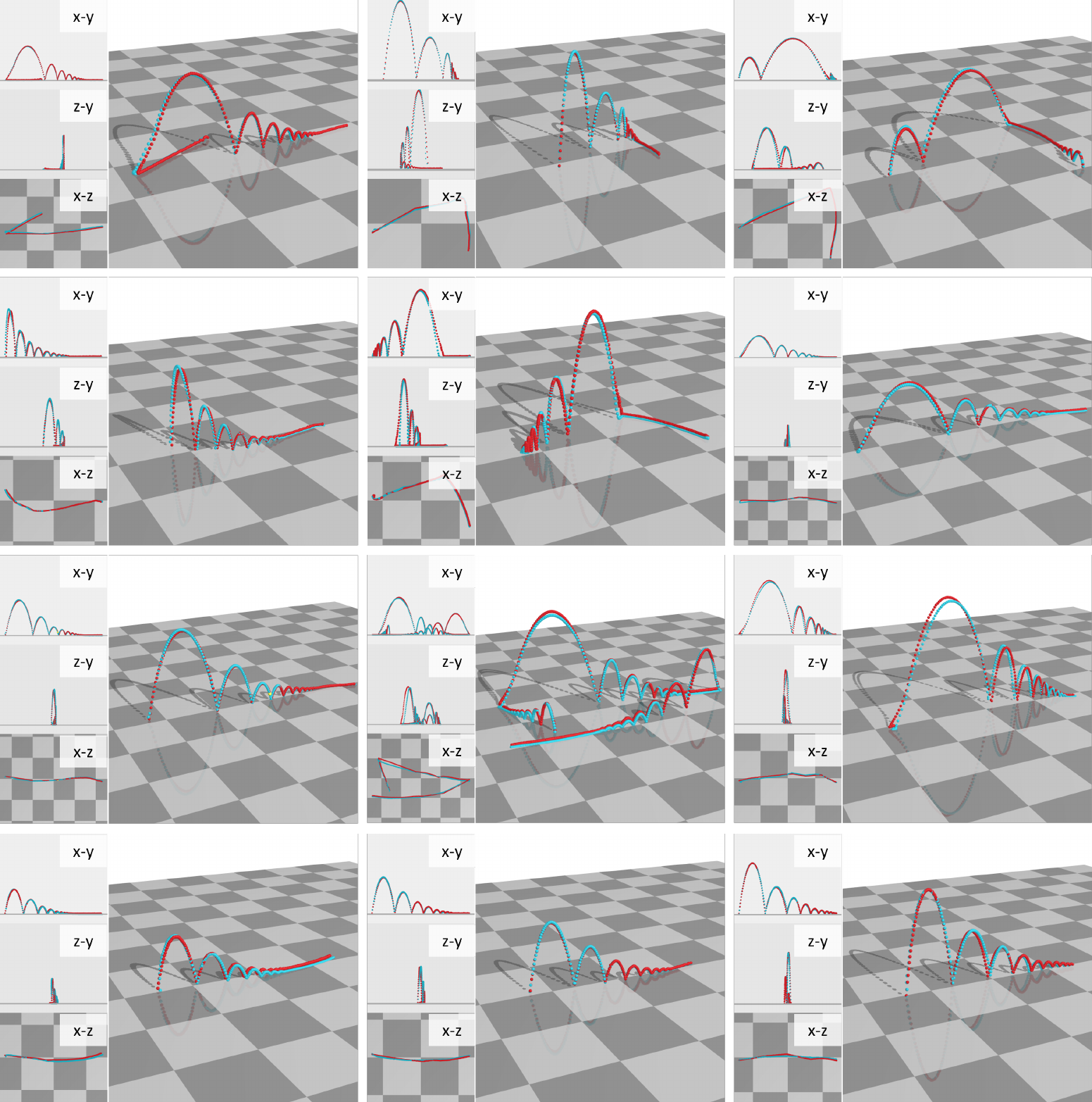}
  \caption{\textbf{Qualitative results on Real Mocap dataset.} Blue: our predictions. Red: ground truth. Each checkerboard block is 75$\times$75 cm$^2$.}
  \label{sm_mocap_all}
\end{figure*}

\begin{figure*}[h]
\centering
  \includegraphics[width=\textwidth]{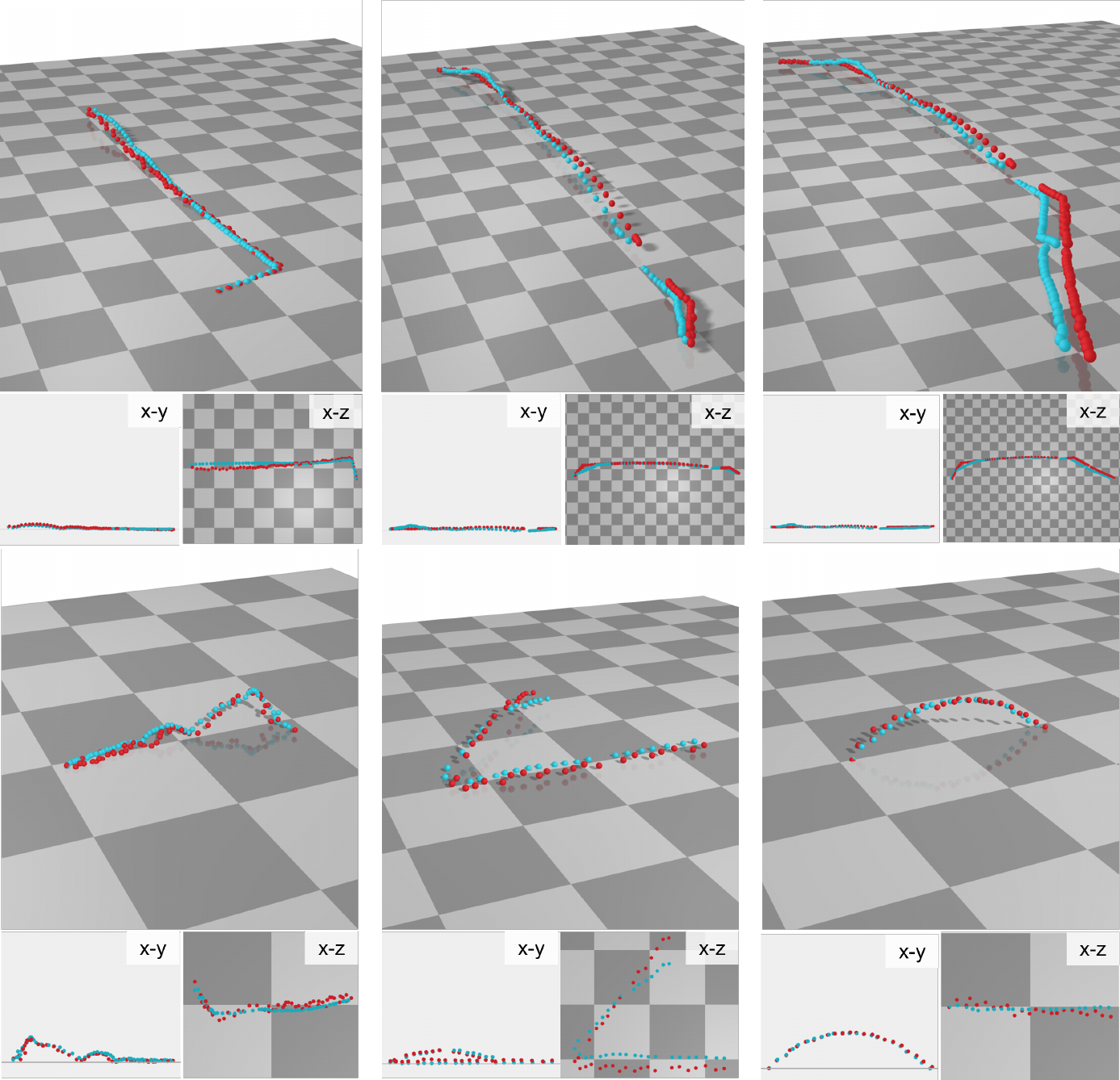}
  \caption{\textbf{Qualitative results on Real IPL dataset.} Blue: our predictions. Red: ground truth. Each checkerboard block is 250$\times$250 cm$^2$.}
  \label{sm_ipl_all}
\end{figure*}

\begin{figure*}[h]
\centering
  \includegraphics[width=\textwidth]{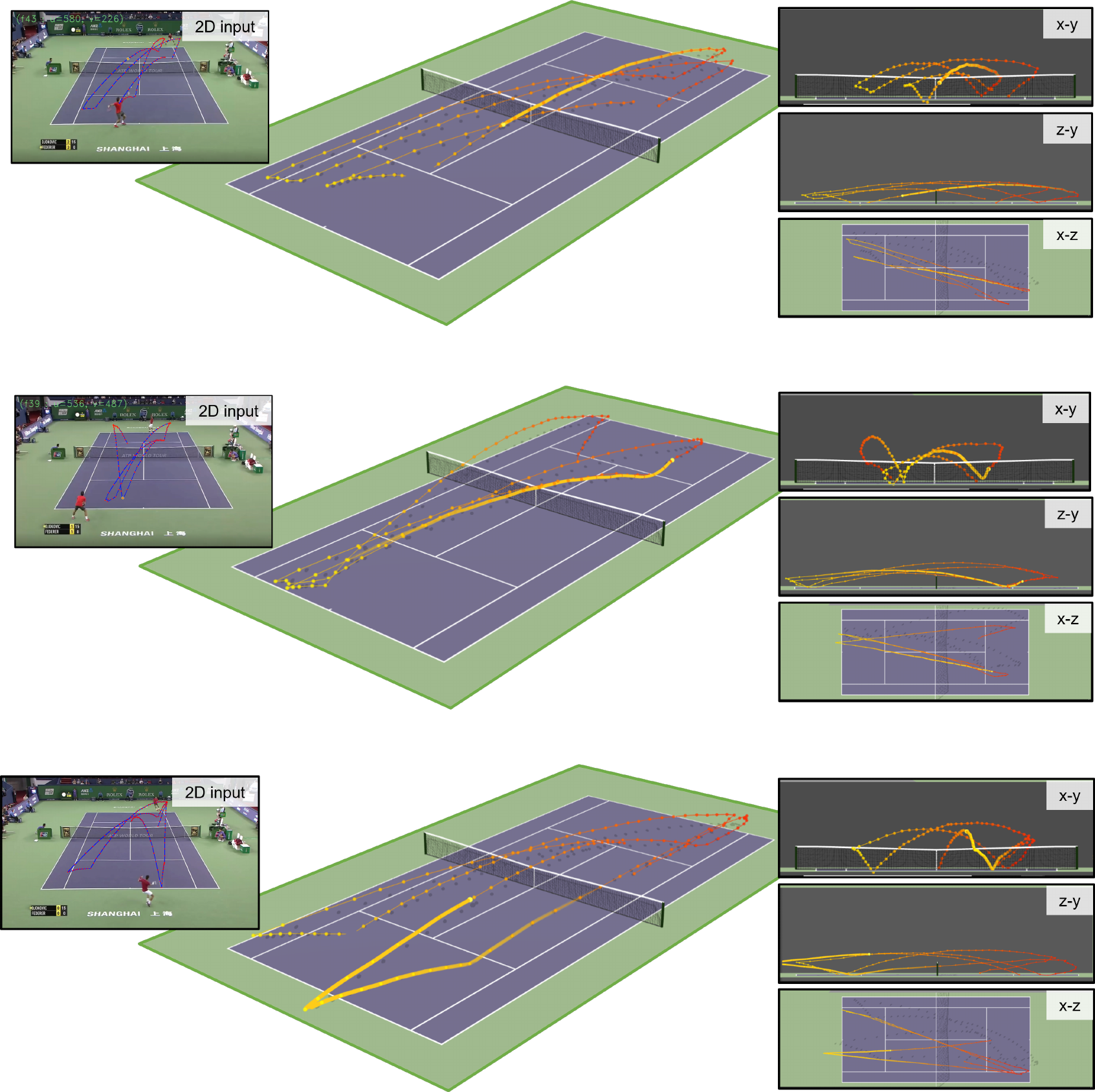}
  \caption{\textbf{Qualitative results on Real Tracknet (Tennis) dataset.}}
  \label{sm_tennis_all}
\end{figure*}

\begin{figure*}[h]
\centering
  \includegraphics[width=0.83\textwidth]{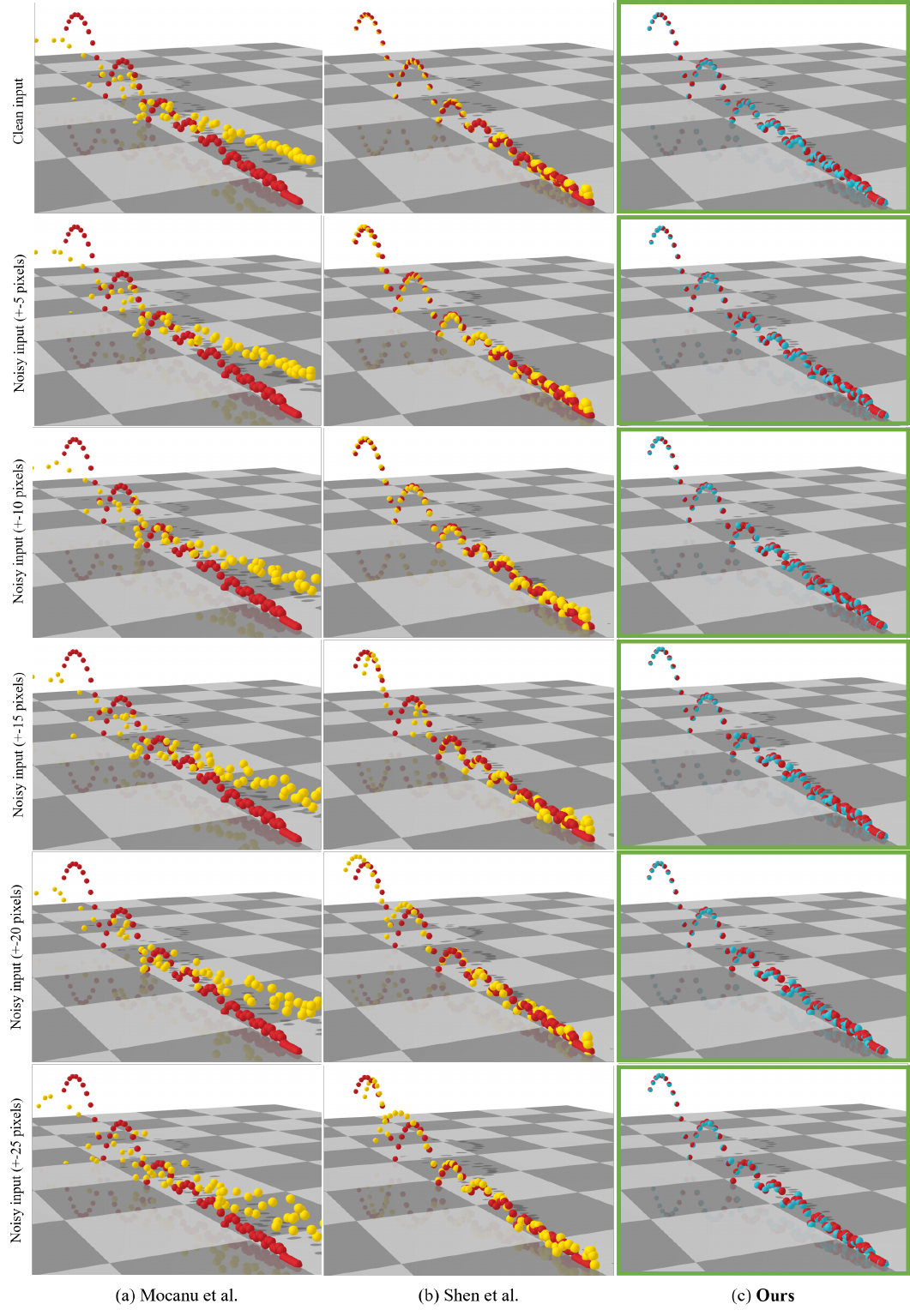}
  \caption{\textbf{State-of-the-art comparison} with a learning-based approach Mocanu et al.\cite{mocanu2017estimating} and a physics-based approach Shen et al.\cite{shen20163d} on a simplified test trajectory that matches their requirements. Each row uses a different noise level. Our predictions are shown in blue, prior work in yellow, and ground truth in red. Each checkerboard block is 50$\times$50 cm$^2$.}
  \label{sm_sota_all}
\end{figure*}

We present additional qualitative results on synthetic datasets of Mocap, IPL, and Tracknet in Figure \ref{sm_synthetic_all}, and on their real counterparts separately in Figures \ref{sm_mocap_all}-\ref{sm_tennis_all}. Lastly, a comparison with the state of the art is shown in Figure \ref{sm_sota_all}.

\section{Failure cases}
\label{app:failurecases}
\begin{figure*}[h]
\centering
  \includegraphics[width=0.99\textwidth]{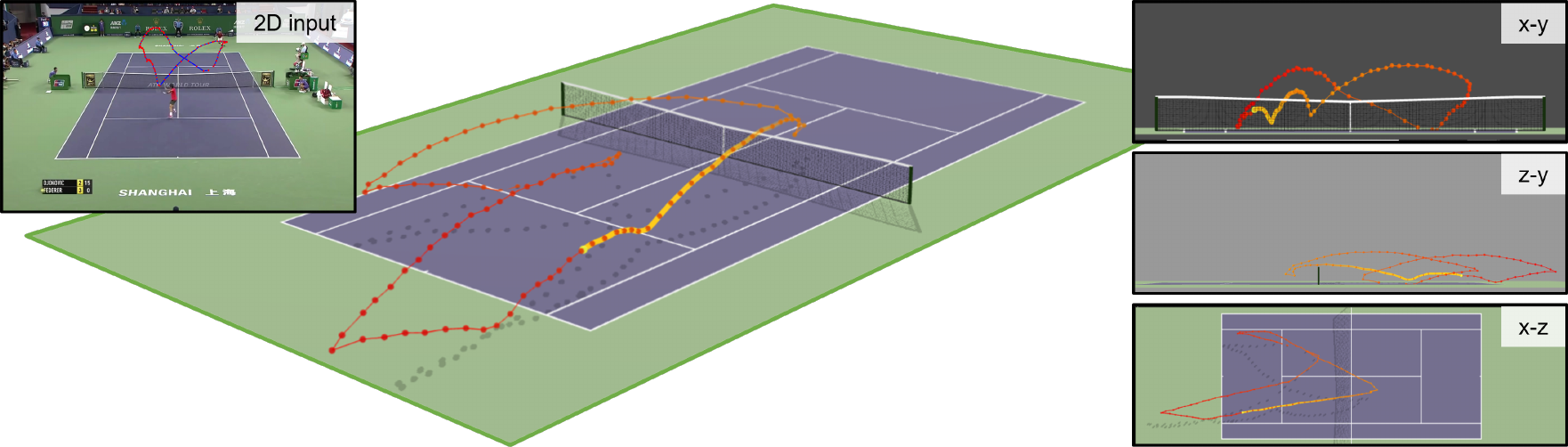}
  \caption{\textbf{Failure cases on Real Tracknet(Tennis) dataset.} This trajectory comes from a volley shot close to the net where the ball bounces right back without hitting the ground, but our prediction shows some slight drop in the ball's height. 
  }
  \label{fc_tennis}

\vspace{1cm}
\centering
  \includegraphics[width=0.96\textwidth]{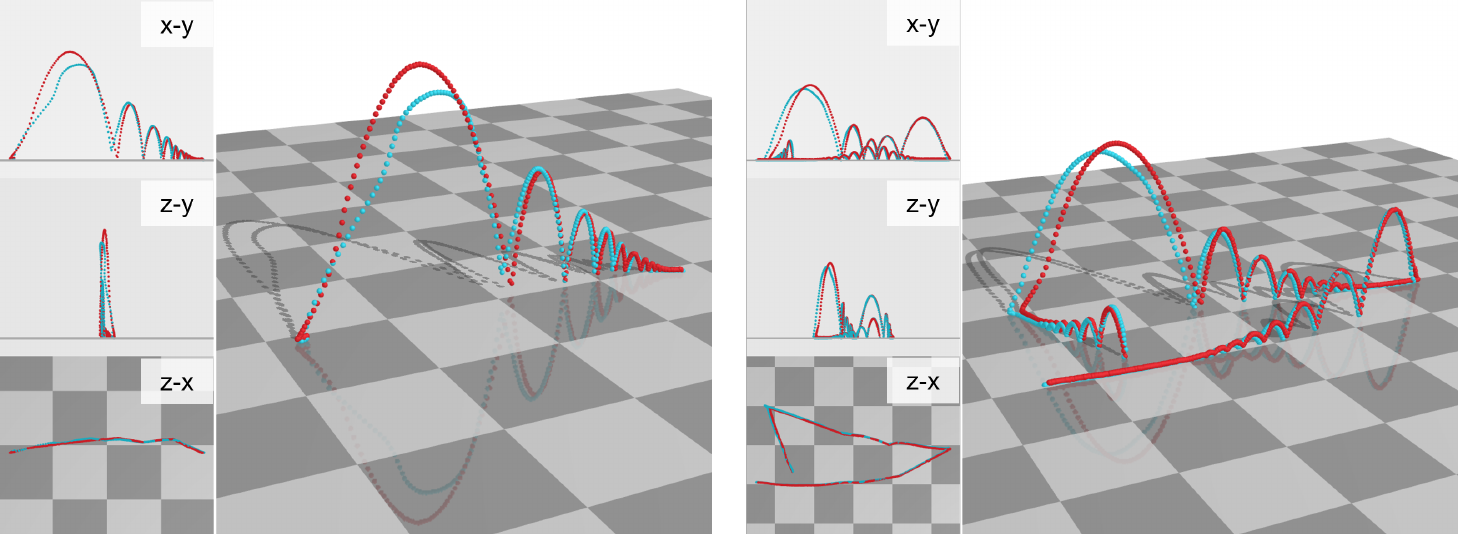}
  \caption{\textbf{Failure cases on Real Mocap dataset.} Blue: our predictions. Red: ground truth. Each checkerboard block is 75$\times$75 cm$^2$.}
  \label{fc_mocap}

\vspace{1cm}
\centering
  \includegraphics[width=0.96\textwidth]{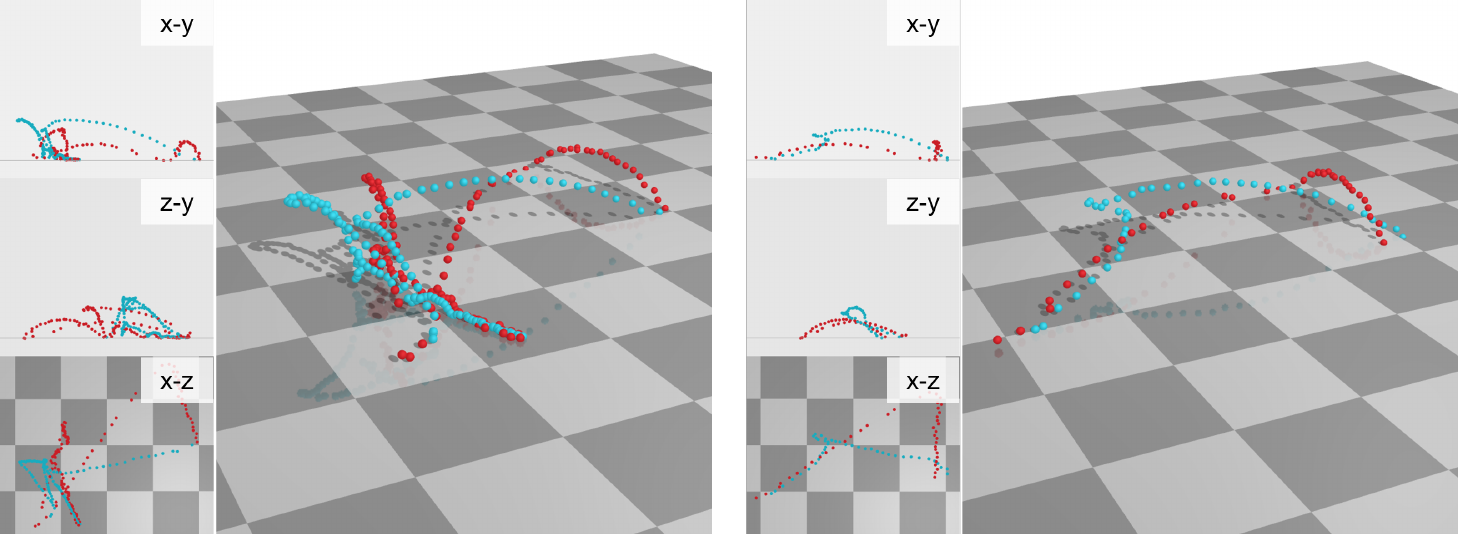}
  \caption{\textbf{Failure cases on Real IPL(soccer) dataset.} When a soccer player chests the ball, the trajectory may look very different from the training trajectories, leading to these errors.
  Blue: our predictions. Red: ground truth. Each checkerboard block is 250$\times$250 cm$^2$.}
  \label{fc_ipl}
\end{figure*}

We observed that our method performs worse on unusual trajectories that are substantially different from the simulated trajectories.
Some rare trajectories in tennis include volley shots (the player returns the ball before it bounces off the ground), or when the player strikes near the net, while in soccer, when the player chests the ball. We show these failure cases on Mocap, Tracknet (Tennis) and IPL datasets in Figure \ref{fc_tennis}, \ref{fc_mocap} and \ref{fc_ipl}. 



  %


\end{document}